\documentclass[preprint,12pt]{elsarticle}

\usepackage{graphicx}%
\usepackage{multirow}%
\usepackage{amsmath,amssymb,amsfonts}%
\usepackage{amsthm}%
\usepackage{mathrsfs}%
\usepackage[title]{appendix}%
\usepackage{xcolor}%
\usepackage{textcomp}%
\usepackage{manyfoot}%
\usepackage{tabularx}%
\usepackage{booktabs}%
\usepackage{algorithm}%
\usepackage{algorithmicx}%
\usepackage{algpseudocode}%
\usepackage{listings}
\usepackage{subcaption}
\usepackage{graphicx}
\usepackage{acronym}
\usepackage{tikz}
\usepackage{comment}
\usepackage{subfig}

\theoremstyle{thmstyleone}%

\begin{document}

\title{A Complete System for Automated 3D Semantic-Geometric Mapping of Corrosion in Industrial Environments}

\author[inst1]{Rui Pimentel de Figueiredo}
\author[inst2]{Stefan Nordborg Eriksen}
\author[inst3]{\\Ignacio Rodriguez}
\author[inst1]{Simon B{\o}gh}

\affiliation[inst1]{organization={Department of Materials and Production, Aalborg University},
city={Aalborg Øst},
postcode={9220}, 
country={Denmark}}
\affiliation[inst2]{organization={Department of Electronic Systems, Aalborg University}, 
city={Aalborg Øst},
postcode={9220}, 
country={Denmark}}
\affiliation[inst3]{organization={Department of Electrical Engineering, University of Oviedo},
city={Gijon},
postcode={33203}, 
country={Spain}}

\begin{abstract}

Corrosion, a naturally occurring process leading to the deterioration of metallic materials, demands diligent detection for quality control and the preservation of metal-based objects, especially within industrial contexts. Traditional techniques for corrosion identification, including ultrasonic testing, radio-graphic testing, and magnetic flux leakage, necessitate the deployment of expensive and bulky equipment on-site for effective data acquisition. An unexplored alternative involves employing lightweight, conventional camera systems, and state-of-the-art computer vision methods for its identification.

In this work, we propose a complete system for semi-automated corrosion identification and mapping in industrial environments. We leverage recent advances in LiDAR-based methods for localization and mapping, with vision-based semantic segmentation deep learning techniques, in order to build semantic-geometric maps of industrial environments. Unlike previous corrosion identification systems available in the literature, our designed multi-modal system is low-cost, portable, semi-autonomous and allows collecting large datasets by untrained personnel. 

A set of experiments in an indoor laboratory environment, demonstrate quantitatively the high accuracy of the employed LiDAR based 3D mapping and localization system, with less then $0.05m$ and $0.02m$ average absolute and relative pose errors. Also, our data-driven semantic segmentation model, achieves around $70\%$ precision when trained with our pixel-wise manually annotated dataset. 
\end{abstract}


\maketitle

\section{Introduction}\label{sec:introduction}

Corrosion is a major problem that degrades metallic surfaces, which are ubiquitous in man-made constructions. Its identification and mitigation have significant socio-economic impacts across various industries, contributing to the maintenance and longevity of infrastructures, such as bridges, pipelines, and buildings, to name a few, reducing the frequency of reconstruction projects and associated costs \cite{koch2002corrosion}.

Early and accurate identification of corrosion allows for timely intervention and maintenance, preventing costly repairs and replacements, reduced downtime in industries such as oil and gas, aerospace, and manufacturing leading to increased productivity and financial savings \cite{koch2017cost}.

Corrosion in metal-made infrastructures can lead to leaks and spills, causing environmental pollution. Timely identification helps prevent such incidents, preserving ecosystems and minimizing the environmental impact \cite{revie2008corrosion}. Also, corrosion-related failures can compromise the safety of structures and equipment. Identifying and addressing corrosion hazards enhances workplace safety and reduces the risk of accidents and injuries, and may lead to lower insurance costs, as the risk of failures and subsequent claims is reduced. Insurers may offer lower premiums for well-maintained and corrosion-free assets \cite{koch2002corrosion}.

Among the leading-edge technologies for identifying it, magnetic flux leakage, ultrasonic testing, and remote visual inspection stand out. Although magnetic flux leakage provides highly accurate solutions, its deployment is expensive, challenging, and necessitates skilled and trained human operators for successful operation. However, the current process of inspecting offshore assets using visual inspection technologies and evaluating where to conduct maintenance and repairs is a highly manual and laborious task. First, inspectors are required to traverse the asset and record and track a large number of images of corroded elements, which are manually labeled in a coarse manner. Also, industrial assets can be multi-story, multi-platform metal structures, naturally resulting in the collection of large datasets, which may be unfeasible to transmit via satellite connections, for instance, in remote offshore locations. Additionally, the nature of industrial assets, as well as various safety and security concerns surrounding them, add challenges to performing the inspection task.

After the data collection procedure, the collected datasets undergo manual review by experts, evaluating the severity of the corrosion, and thus which structures require maintenance, and for which maintenance can be postponed. Some structures are more critical than others to maintain, for the sake of asset operation and safety. As part of this review, most of the time is spent resolving where a given image was recorded, and identifying the specific location of the corroded structure. This process involves the development of maintenance work packages. Within these packages, structures identified as critically corroded are designated for maintenance, repair, or replacement. These tasks may extend over several months, whether carried out on-site or, in the case of mobile offshore platforms, at a dry dock.

To alleviate some of the challenges posed by the manual nature of the current process, we envision a data collection device with the following general properties:

\begin{itemize}
    \item \textbf{Portability}: Inspectors should be able to easily carry and operate the data collection device in industrial assets during long periods.
    \item \textbf{Accuracy}: High precision imaging sensors for accurate 3D semantic mapping of corroded assets.
    \item \textbf{Autonomy}: Intelligent software that should be able to perform mapping, localization and semantic localization and categorization of industrial assets, from sensory data. However, the software may not strictly run online, on the data collection device, but instead offline in a more powerful computational device.
\end{itemize}

Taking into account the previous properties, we leverage current advances in vision-based semantic segmentation and simultaneous localization and mapping (SLAM) approaches, and the availability of low-cost consumer grade LiDAR and visual camera systems, to propose a complete geometric-semantic mapping and localization system. We combine highly accurate 3D point cloud data provided by LiDAR, with monocular color vision for robust and precise localization of corrosion spots in 3D.

The rest of this article is organized as follows. First, in Section~\ref{sec:soa}, we review the state-of-the-art technologies for positioning and corrosion identification, with a focus on vision-based 3D localization and mapping and semantic segmentation techniques. Second, in Section~\ref{sec:system} we describe in detail the proposed system, including design and sensor choices, as well as algorithmic choices. Then, in Section~\ref{sec:experiments} we validate our system and demonstrate its applicability by reporting the results from a set of experiments performed both in an indoor laboratory scenario and in an outdoor offshore environment. Finally, we draw our main conclusions and propose extensions for future work in Section~\ref{sec:conclusion}.

\section{Related work}\label{sec:soa}

In this section we overview the state-of-the-art methods utilized in our approach, including existing technologies for positioning, corrosion detection, vision-based sensor calibration, localization and mapping, and semantic segmentation techniques.

\subsection{Positioning Systems}
Multiple positioning technologies exist and have been reported in the literature for multiple purposes. With a focus on offshore positioning cases, industrial positioning, emergency systems, and positioning systems for autonomous systems, the following was concluded. 

Technologies such as Ultra-Wide Band (UWB)~\cite{uwb}, Ultra-sound~\cite{ultrasound}, Wi-Fi~\cite{wifi_fingerprinting}, and Bluetooth Low-Energy (BLE)~\cite{ble} can work in indoor settings and are simple to deploy as end user devices are typically based on tags or small battery-power devices. However, these technologies present also some drawbacks. They normally require a dedicated and complex infrastructure deployment with anchors/routers, wires, and centralized control computers; as well as an accurate calibration process. While in typical industrial settings such as factory halls these type of deployments might be feasible, the situation in offshore is more complicated (e.g., in oil/gas platforms). In general, these technologies require clear line-of-sight between infrastructure and localization devices to guarantee a correct operation (i.e., at least three visible routers/anchors at all times to be able to perform triangulation), which is difficult in both industrial and offshore scenarios. Moreover, localization based on these technologies does typically not provide heading/orientation information and does not perform well when the localization devices are close to large metal surfaces. The accuracy of these systems in industrial settings typically range from approximately 20~cm for UWB (with some sporadic deviations to more than 1~m in certain conditions), to approximately 1.5~m for Wi-Fi and BLE.

Technologies based on satellite systems such as Global Navigation Satellite System~(GNSS)/Real Time Kinematic~(RTK)~\cite{gnss}, and Differential Global Positioning Systems~(DGPS)~\cite{dgps} can offer up to cm-level accuracy in industrial and offshore environments. However, their applicability is limited to areas with a full or partial view of the sky, hence inapplicable in indoor settings~\cite{limitationgnss}. In general, these technologies rely on no infrastructure (e.g., GNSS, where only an end user device is required) or simple infrastructure (e.g., RTK, DGPS), where a single unit or just a few independent units are typically needed for automatic systems calibration/correction.

As an alternative, it is possible to use novel technologies or combinations of technologies that do not rely on external infrastructure~\cite{intsolut}. Here, recent advances in Computer Vision~(CV)~\cite{machinevision} methods for Simultaneous Localization and Mapping (SLAM)~\cite{ar}, and Visual-Inertial Odometry (VIO)~\cite{vio} are at hand, allowing for simultaneous high-precision, real-time and low-cost state-estimation (i.e. tracking), provisioning of 6 degrees of freedom (6 DoF) position and orientation information and mapping. Of course, all these benefits have some associated limitations/drawbacks that need to be overcome such as the unavailability of unified commercial solutions, and more complicated integration setups. Other challenges may include occasional poor performance in featureless areas, and the potential need for high computational resources and stability requirements for mobile setups~\cite{inspection_techs}. Sensor fusion strategies (e.g., integration with Inertial Measurement Units (IMU)) might be of utility to  overcome some of the said challenges. 

Therefore, to make our proposed solution as universal as possible, relying on external infrastructure or specific conditions is avoided, making it compatible with most of the potential industrial and offshore scenarios. Hence, the last group of novel technologies, and its combination, will be explored.

\subsection{Technologies for corrosion identification}

Identifying corrosion relies on diverse technologies \cite{roberge2007corrosion} capable of detecting, quantifying, and characterizing corrosion processes in various materials and environments. Commonly employed methods include Ultrasonic Testing (UT) \cite{kapoor2016parameters}, which utilizes high-frequency sound waves for internal and surface defect detection in industries like oil and gas, aerospace, and manufacturing. Eddy Current Testing (ECT) \cite{sophian2017pulsed} is a method that utilizes electromagnetic induction, being particularly useful for inspecting non-ferrous metals and detecting corrosion under protective coatings. Radiographic Testing (RT) \cite{bortak2002guide} involves X-rays or gamma rays to inspect internal structures, commonly applied in aerospace, construction, and automotive industries.

Electrochemical Techniques (ET) \cite{kelly2002electrochemical}, encompassing methods like impedance spectroscopy and potentiodynamic polarization, study the electrochemical behavior of metals in corrosive environments, providing vital information on corrosion rates and protection effectiveness. Infrared Thermography (IRT) \cite{doshvarpassand2019overview}, a non-contact method using infrared cameras, is beneficial for inspecting large areas in industries such as building inspection, aerospace, and electrical utilities. Visual Inspection \cite{choi2005morphological}, a basic method involving the visual examination of surfaces, is often used alongside other testing methods to assess corrosion severity. Additional technologies like magnetic particle testing \cite{lovejoy1993magnetic}, acoustic emission testing \cite{zaki2015non}, and scanning electron microscopy \cite{aharinejad2012microvascular} also contribute significantly to corrosion identification.

These technologies play a crucial role in preventing material degradation, structural failure, and costly repairs. Our proposed pipeline employs a vision-based deep-learning method, leveraging low-cost camera sensors for accurate and easily deployable results in corrosion assessment.

\subsection{Camera and LiDAR calibration}

Camera, LiDAR and inertial (i.e. IMU) calibration is a crucial step in sensor fusion for robotics, autonomous vehicles, and augmented reality applications \cite{debeunne2020review}. This process ensures accurate alignment and synchronization between these sensors. Calibration enables the transformation of measurements from different sensors into a common reference frame, allowing seamless integration of data for robust perception and navigation.

Camera calibration involves determining the intrinsic and extrinsic parameters of a camera, which include focal length, principal point, distortion coefficients, and the transformation (rotation and translation) between the camera and the world coordinate system. The calibration is often performed using a calibration target with known geometric features (typically a checkerboard pattern with known dimensions), and the parameters are estimated by minimizing the re-projection error \cite{zhang2000flexible}. 

A set of algorithms for the calibration of various camera sensors, commonly used in robotics applications (e.g. for precise sensor fusion and localization), and can be found in a popular open-source toolbox named Kalibr \cite{rehder2016extending}. These algorithms are not limited to single camera systems and may be used to calibrate multiple cameras (stereo), of various types, and IMUs (Inertial Measurement Units).

LiDAR-camera calibration involves aligning LiDAR data, with camera to ensure accurate fusion of LiDAR point clouds with camera. The calibration process determines the transformation matrices between the LiDAR and camera  coordinate systems \cite{mirzaei20123d}. Calibration, in this case, is typically performed by capturing synchronized data from both sensors while the system undergoes controlled motion.

In all cases, optimization techniques, such as nonlinear least squares, are used to estimate the transformation matrices holding the sensors' intrinsic and extrinsic parameters.




\subsection{LiDAR-based localization and mapping}

LiDAR-based SLAM is an advanced technology that combines data from LiDAR sensors and optionally inertial measurement units (IMUs) to achieve real-time localization and mapping in dynamic environments. LiDAR sensors employ laser beams to measure distances and create detailed three-dimensional (3D) maps of the surroundings, while IMUs, equipped with accelerometers and gyroscopes, capture information about the platform's acceleration and angular rate. The integration of LiDAR and IMU data enhances the accuracy and robustness of the system, especially in scenarios where Global Navigation Satellite Systems (GNSS) may be unreliable, such as indoor or urban environments.

Approaches for LiDAR based SLAM, involve estimating the precise position and orientation of the sensor apparatus in relation to its surroundings, while building a map representation of the environment (typically a point cloud),  without prior knowledge of the surroundings. This is achieved through continuous fusion and optimization of LiDAR and, if available, IMU measurements, enabling the system to maintain an accurate and up-to-date understanding of its pose. The integration of these sensors facilitates overcoming challenges like dynamic movements, changes in orientation, and variations in the environment. Loop closure detection is crucial technique employed in SLAM for correcting accumulated errors in both localization and mapping, and typically occurs when the sensing agent revisits a previously seem location \cite{arshad2021role}.

When a map of the environment is known a priori, SLAM simplifies to a localization problem, in which the goal is to estimate the sensor apparatus pose with respect to the map coordinate system. LiDAR based SLAM and localization systems are crucial for various applications, including robotics, autonomous vehicles, and augmented reality, where high-precision spatial awareness and real-time localization are paramount. Researchers and engineers continue to refine and develop LiDAR SLAM algorithms to enhance performance, robustness, and adaptability across diverse and challenging scenarios. We overview the most relevant state-of-the-art approaches below.

\subsubsection{LiDAR-based SLAM approaches}
A similar work to ours is the one of \cite{koide2019portable} that proposes a portable LiDAR system for long-term and wide-area people behavior tracking. The authors utilize an optimization-based Simultaneous Localization and Mapping (Graph-SLAM) \cite{thrun2006graph} approach for mapping the environment, while estimating the system's pose and concurrently tracking target individuals. The system operates in two distinct phases: 1) offline environmental mapping and 2) online sensor localization and people detection/tracking. During the offline mapping phase, a comprehensive 3D environmental map covering the entire measurement area is generated. The mapping process employs a Graph-SLAM approach. To address accumulated rotational errors in scan matching, ground plane and GPS position constraints are introduced for indoor and outdoor environments, respectively. The proposed system thus employs a sophisticated mapping strategy to create accurate and comprehensive environmental 3D point cloud representations.

A recent approach entitled Fast-LIO \cite{fastLio21}, proposed a LiDAR-Inertial Odometry (LIO) framework featuring a Tightly-Coupled Iterated Kalman Filter for robust and efficient motion estimation. The pipeline integraties LiDAR and inertial sensor measurements through a Kalman filter framework. Fast-LIO is designed for real-time applications, such as robotics and autonomous navigation, offering fast and accurate odometry estimation. The key contributions of the approach lie in the technical advancements of the Tightly-Coupled Iterated Kalman Filter, enhancing the reliability and speed of LiDAR-Inertial Odometry for precise motion tracking. In a second iteration \cite{xu2022fast}, the authors propose two new techniques to further improve the performance of the previous algorithms. The first involves direct registration of raw points to the map, eliminating the need for feature extraction. This approach enhances accuracy by exploiting subtle environmental features and is adaptable to various LiDARs. The second novelty lies in maintaining a map using the ikd-Tree, a new incremental k-d tree data structure. This structure facilitates incremental updates, point insertion/deletion, and dynamic re-balancing, demonstrating superior overall performance compared to existing dynamic data structures while supporting down-sampling on the tree.

\subsubsection{LiDAR-based localization approaches}
In the work of \cite{akai2023efficient}, the authors propose a 3D Monte Carlo approach for localization in dynamic environments in the context of automated driving, utilizing efficient distance field representations. Their approach enhances accuracy in estimating the vehicle's position, especially in scenarios with dynamic elements. The study employs Monte Carlo methods for three-dimensional localization while optimizing the representation of the surrounding environment through efficient distance fields. The research has implications for improving the precision of autonomous vehicles navigating through dynamic settings.

In the work of \cite{amcl3d2017}, the authors propose a multi-sensor three-dimensional Monte Carlo localization method designed for long-term aerial robot navigation. The approach integrates information from multiple sensors to enhance the precision and reliability of localization. Utilizing Monte Carlo methods, the system estimates the aerial robot's three-dimensional position over time. The multi-sensor setup aims to address the challenges associated with long-term navigation, offering improved adaptability and robustness. The article discusses the technical details of the proposed method, emphasizing its applicability for sustained and accurate aerial robot navigation in dynamic environments.

In the work of \cite{vizzo2023ral}, the authors advocate that for the utilization of a simplified and refined Point-to-Point Iterative Closest Point (ICP) registration method, referred to as KISS-ICP. The authors assert that when implemented correctly, this approach offers simplicity, accuracy, and robustness in the context of point cloud registration. The article emphasizes the importance of proper execution to achieve optimal results. KISS-ICP is presented as an efficient and effective solution for point-to-point registration tasks, contributing to improved performance in various applications such as 3D mapping, computer vision, and robotics.

In \cite{akai_arxiv2023_mcl3d}, the authors propose an effective approach for 3D LiDAR-based Monte Carlo based localization. The method involves the fusion of measurement models optimized through importance sampling, aiming to enhance the efficiency and accuracy of the localization process. The article presents a detailed analysis of the proposed solution, highlighting its advantages in terms of computational efficiency and robustness in handling complex 3D environments. The fusion of optimized measurement models contributes to improved localization performance, making the approach a valuable contribution to the field of autonomous navigation and robotics using 3D LiDAR sensor data.

\subsection{Image-based semantic segmentation}
The comprehension of scenes through vision-based analysis holds paramount  significance across diverse domains, including robotics, manufacturing \cite{visionCorrosionIST2023}, medical imaging \cite{mlBio2020}, inspection  \cite{jimaging8030062,9551637,9635986, ba251ca5d8704ecbb39cb303f63473ab}, and surveillance \cite{gruosso2021human}. Scene understanding encompasses varied tasks, including image classification, object detection, and semantic segmentation.  Vision-based semantic segmentation, addresses the intricate challenge of assigning object class labels to individual pixels within images.

This section undertakes a comprehensive revision and explanation of primary techniques aimed at addressing the aforementioned semantic segmentation challenge. While image classification and object detection respectively deal with classifying images and localizing regions of interest (bounding boxes), semantic segmentation tackles the more intricate task of assigning a class label to each pixel in an image. Approaches documented in the literature can be classified into two distinct paradigms: model-based and data-driven.

Classical computer vision methodologies are based on theoretically principled methods striving to analytically resolve geometric and physical aspects of image formation. In contrast, data-driven methodologies seek to learn statistical properties directly from visual data using machine learning techniques. The ascendancy of deep learning, coupled with the availability of extensive publicly annotated datasets, e.g., Ms COCO \cite{Lin2014MicrosoftCC}, CityScapes \cite{cordts2016cityscapes}, AED20K \cite{zheng2021rethinking}, to name a few, has resulted in data-driven approaches consistently outperforming their model-based counterparts, particularly in addressing increasingly complex tasks.

One of the initial efficacious forays into deep learning-based semantic segmentation was Mask R-CNN \cite{he2017mask}. The architecture exhibits conceptual simplicity, comprising a Convolutional Neural Network (CNN) backbone for robust feature extraction, succeeded by a meticulously optimized Region Proposal Network (RPN) tasked with generating candidate regions of interest. Subsequently, three parallel branches undertake the tasks of classification, bounding box regression, and pixel-level mask predictions. Mask R-CNN, along with subsequent architectures sharing analogous design principles, has demonstrated state-of-the-art performance across diverse semantic segmentation datasets, notably showcasing excellence on the Microsoft COCO dataset \cite{lin2014microsoft}.

U-net, a fully convolutional neural network introduced in \cite{unet2015}, represents a step forward from previous deep learning  based semantic segmentation approaches. This architecture hinges on the novel concept of substituting fully connected layers with up-sampling layers, thereby enabling precise pixel-level predictions. More specifically, U-net adopts an encoder-decoder architecture devoid of fully connected layers. The CNN encoder, operational along the contracting path, strategically downscale the input image to a low-dimensional feature space. Simultaneously, the expansive path of the decoder utilizes de-convolutional layers to up-sample the feature space. U-net has achieved notable success, particularly in the realm of biomedical imaging applications, where it has proven effective in the segmentation of tumor cells. Due to its top performance and popularity, and based on our previous experimental work \cite{visionCorrosionIST2023}, we utilize U-net model as our corrosion semantic segmentation approach.

\section{Methodologies} \label{sec:system}
In this section we describe in the detail the design choices behind the proposed portable hand-held LiDAR-inertial camera and data collection system, and our approaches for 3D mapping and localization of corrosion in industrial environments.

\subsection{Problem formulation}
Offshore assets are not the only locations where corrosion must be managed, but provides an extreme case, where even gaining access to the site is prohibitive. The person conducting the inspection is also unlikely to be the same person evaluating the severity of corrosion. As such, the latter cannot to the same degree rely on prior knowledge of the site when reviewing the dataset.

Considering the presented issues with offshore corrosion management, the overall properties of the envisioned system, while simultaneously acknowledging that such a system may prove beneficial in onshore settings as well, we present a portable LiDAR-inertial camera system that can automatically localize and identify corroded structures, in industrial sites.

In particular, we address several key challenges:

\begin{itemize}
    \item \textbf{Perception and data collection system design:} To support corrosion inspections at various environments, without the need for any external permanent fixtures, a portable platform for hosting sensors and computational resources is required. The platform should be lightweight, portable, and allow recording RGB and point cloud data streams in outdoor environments for prolonged periods of time. We envision a handheld sensor apparatus containing a camera and LiDAR, and a backpack computer and power-system.
    \item \textbf{Sensor calibration:} With the purpose of projecting image-based data accurately to the LiDAR point cloud, we need to find the pose of the \textit{camera system}, relative to the LiDAR, i.e., the extrinsic camera matrix parameters, encoding a transformation between the camera and the LiDAR.  This way, one can accurately map RGB and semantic information to LiDAR point clouds, in order to build detailed colored point clouds, representing the surrounding environment.
    \item \textbf{3D localization and mapping:} Offshore assets can be geometrically complex and multiple stories tall, and images are also likely to be captured from varied poses. As such, the localization system must be able to operate in 3D, i.e., a detailed map of the environment is required for localization, and must be generated using a LiDAR and a camera system.
\item \textbf{Corrosion detection:} images may contain numerous small spots of corrosion that can be missed, and some images may feature corrosion runoff or other visual blemishes that are not relevant on their own. Corrosion is also quite varied, depending on the environment and type of metal. A well-defined semantic segmentation system is needed for this task. 
\end{itemize}

\begin{figure*}
\centering
\includegraphics[width=0.99\linewidth]{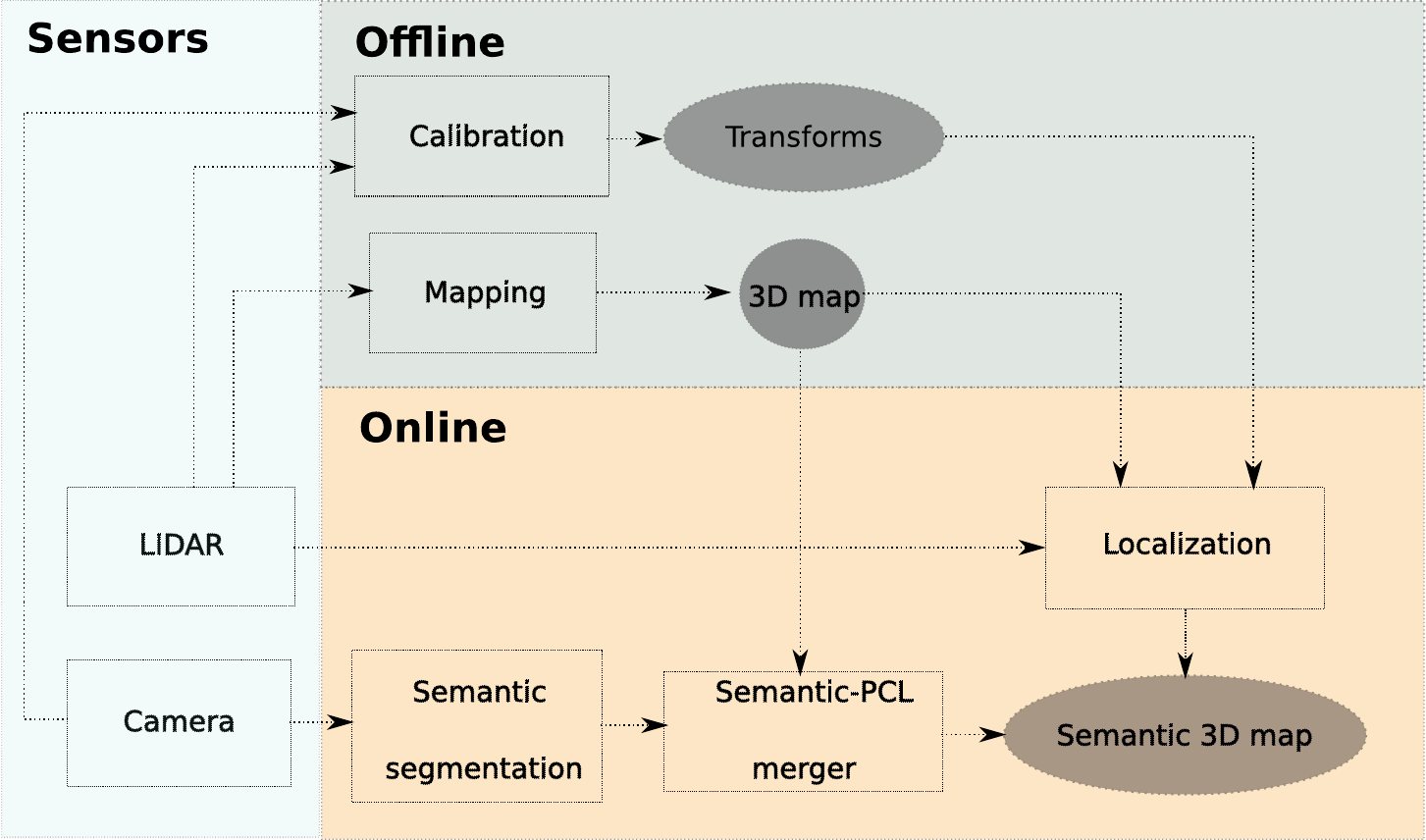}
\caption{Proposed modular pipeline for semantic-geometric mapping of corrosion in metallic surfaces.}
\label{fig:pipeline}
\end{figure*}

In the rest of this section we describe in detail our proposed solution that addresses the former challenges, through the use of state-of-the-art SLAM and image-based semantic segmentation techniques, for geometric-semantic mapping of corrosion in industrial environments (see Figure \ref{fig:pipeline}).

\subsection{The Design of a Portable Corrosion Identification System}

\begin{figure}
    \centering
    \begin{subfigure}[t]{0.9\textwidth}
        \centering
        \includegraphics[width=\textwidth]{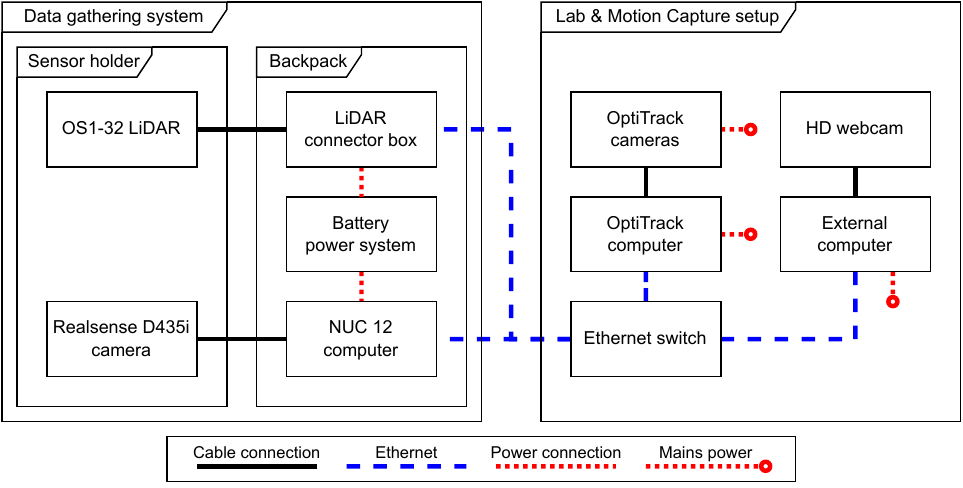}
        \caption{Diagram of envisioned system.}
        \label{fig:system_diagram_vision}
    \end{subfigure}
    \begin{subfigure}[t]{0.35\textwidth}
        \centering
        \includegraphics[width=\textwidth]{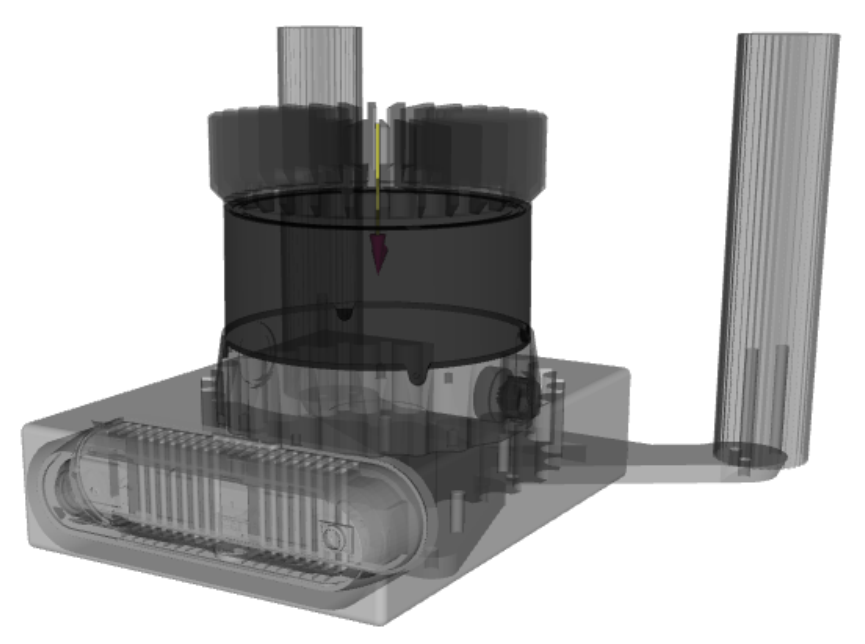}
        \caption{Sensor-holder CAD model.}
        \label{fig:sensor_holder_cad}
    \end{subfigure}
    \begin{subfigure}[t]{0.98\textwidth}
        \centering
        \includegraphics[width=0.62\textwidth]{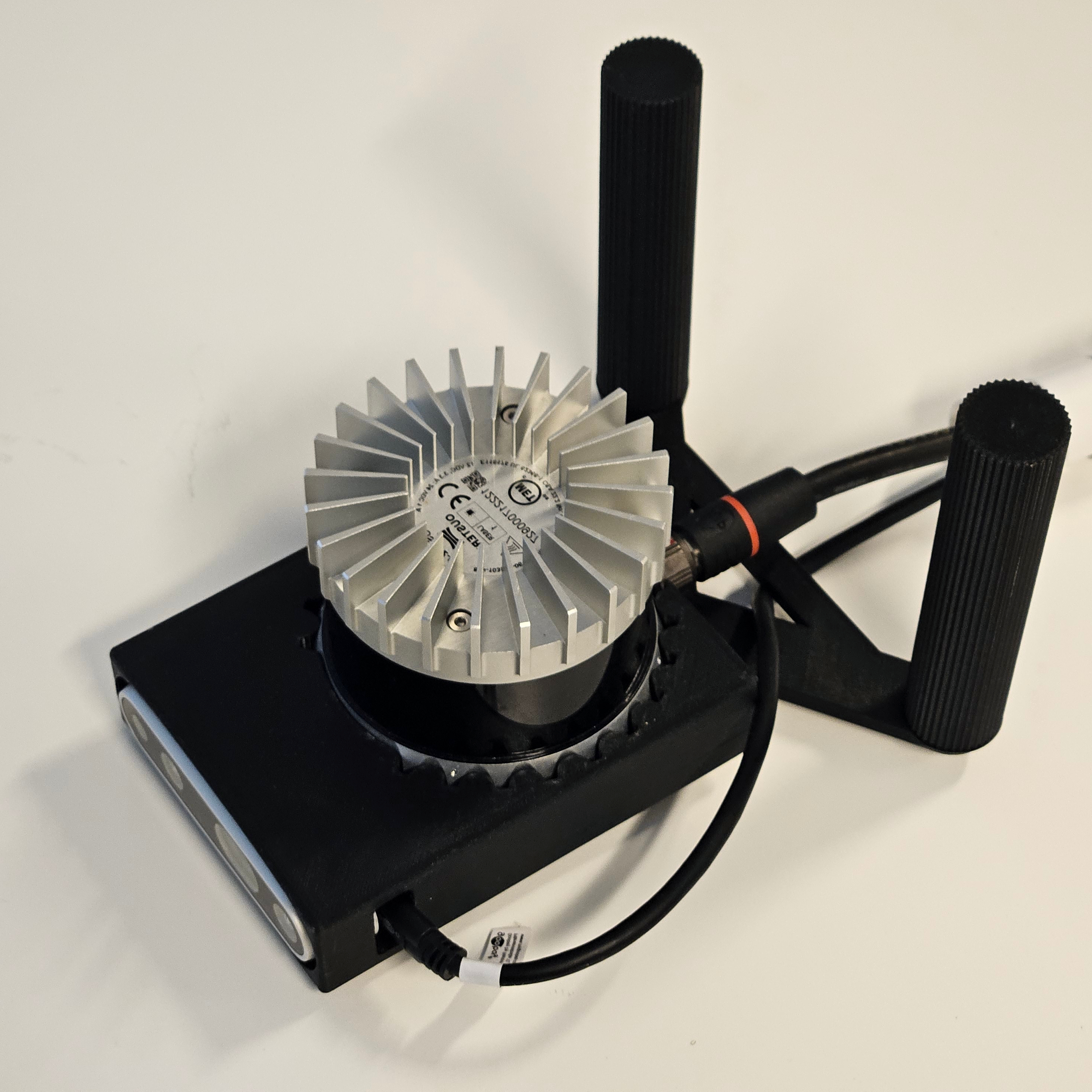}
        \includegraphics[width=0.3\textwidth]{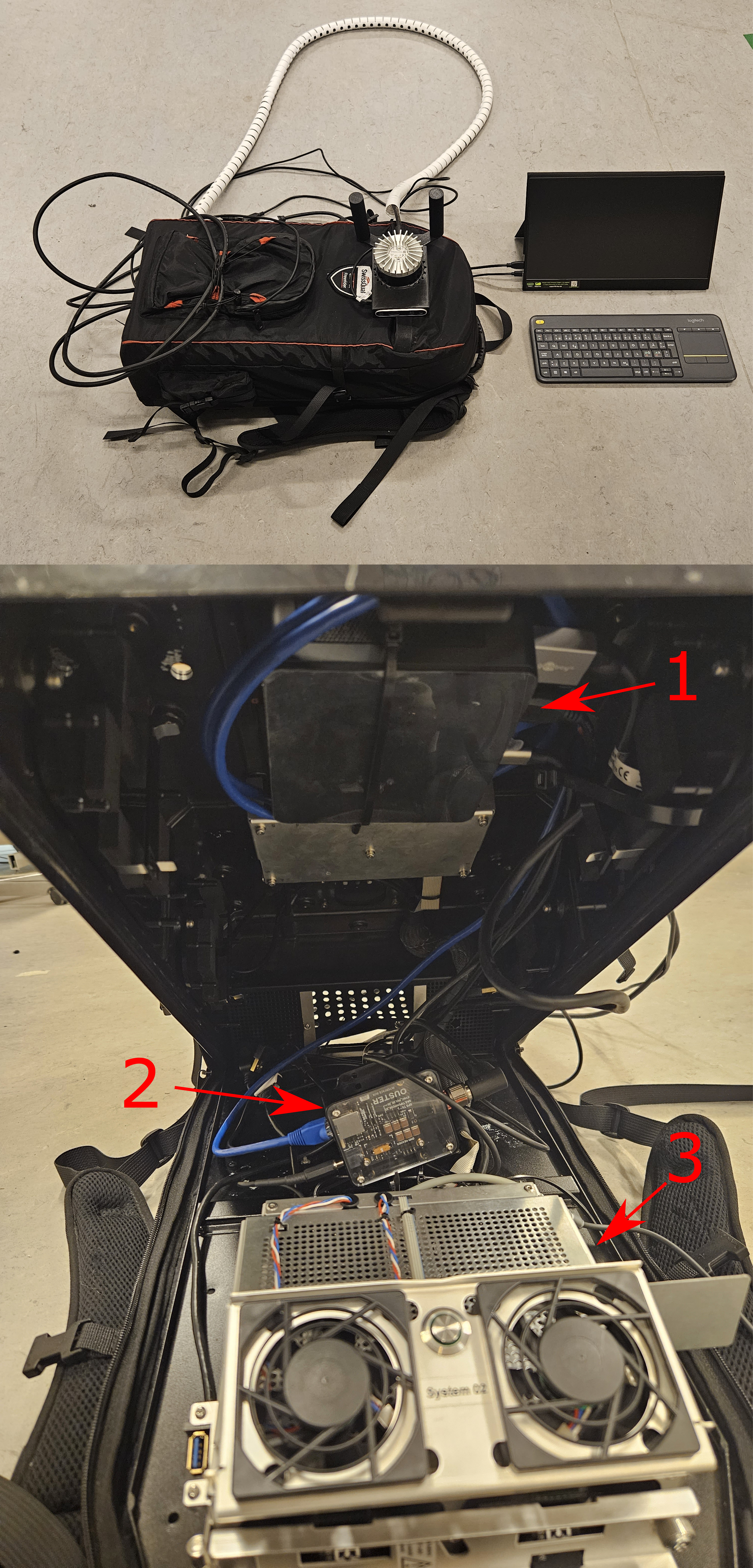}
        \caption{Left: Assembled sensor-holder apparatus. Top right: Full portable system, with optional portable monitor and wireless keyboard. Bottom right: Internals of the backpack, showing computer (1), LiDAR connector box (2) and battery power system (3).}
        \label{fig:backpac-system}
    \end{subfigure}    
    \caption{Detailed design view of our portable data capture system for 3D semantic-geometric mapping.}
    \label{fig:system_design}
\end{figure}

The design of a portable corrosion identification system leveraging LiDAR-inertial camera sensors represents a novel approach in corrosion assessment technology. 

Our proposed system integrates a LiDAR and inertial measurement units (IMUs) for capturing motion and orientation data, for precise 3D localization and mapping, and camera sensors for visual inspection of corrosion spots on metallic surfaces. The fusion of these technologies enables a comprehensive and accurate evaluation of corrosion in diverse environments, in particular, industrial environments. The LiDAR-inertial camera sensors provide the system with the capability to create high-resolution 3D models of the environment, and detailed corrosion mapping.

The portability aspect ensures the system's adaptability for field applications, facilitating on-site inspections in real-world conditions. The design prioritizes the development of a lightweight and user-friendly system that combines the strengths of LiDAR, inertial, and camera sensors to deliver reliable corrosion identification in a portable and efficient manner, catering to the evolving needs of corrosion inspection in various industries.

\subsection{LiDAR-Camera Geometry}
\label{sec:lidar_inertial_camera_model}
In order to be able to map 2D camera image pixels $(u,v)\in \mathbb{N}^2$ to LiDAR 3D coordinates $(x_l,y_l,z_l)\in \mathbb{R}^3$ and vice-versa, one needs to know the  intrinsic parameters and the relative transformation between both sensors. This mapping can be expressed in linear homogeneous coordinates, according to

\begin{align*}
\begin{bmatrix}
u\\
v\\
1\end{bmatrix}=\mathbf{K} 
\mathbf{T}_{l\rightarrow c}\begin{bmatrix}
x_{l}\\
y_{l}\\
z_{l}\\
1\end{bmatrix}
\end{align*}
where $\mathbf{K}$ denotes the intrinsic camera parameters matrix, encoding the transformation from the camera coordinate system $\mathcal{C}$ to the pixel coordinate system $\mathcal{I}$, and $\mathbf{T}_{l\rightarrow c}
$ the transformation from the LiDAR coordinate system $\mathcal{L}$ to the camera coordinate system $\mathcal{C}$. Transformations and 2D and 3D coordinates are, through the rest of the article, expressed in homogeneous form.

\subsubsection{Intrinsic camera parameters} To know exactly how points in the scene map to the image plane for a given camera, it is necessary to find the intrinsic camera matrix $\mathbf{K}$,
    \begin{equation}
 \mathbf{K}={
\begin{bmatrix}\alpha _{x}&\gamma &u_{0} & 0\\0&\alpha _{y}&v_{0}&0\\0&0&1&0
\end{bmatrix}}
    \end{equation}
where $\alpha_x=f m_x$ and $\alpha_y=f m_y$, represent the focal distance in pixels, with $f$ denoting the focal distance in camera (metric) coordinates, and $m_x$ and $m_y$ the inverses of the width and height of a pixel in the image plane. 

In order to model image distortion due to lens characteristics, we use the Zhang model \cite{zhang2000flexible}, a widely used lens distortion model in camera calibration. The model introduces rectification coefficients to correct for the non-ideal behavior of lenses, especially radial distortions, incorporating both radial and tangential distortion effects in a camera lens. The model expresses the distorted image coordinates $(u',v')$
 in terms of the undistorted coordinates $(u,v)$ using the following polynomial forms,
\begin{align*}
u'=u\left (1 + k_1  r^2 + k_2 r^4 k_3 r^6 \right ) + 2p_1uv+p_2(r^2+2u^2)\\
v'=v\left (1 + k_1  r^2 + k_2 r^4 k_3 r^6 \right ) + p_1(r^2 + 2v^2) + 2p_2uv 
\end{align*}
where $k_1,k_2,k_3$ denote the radial distortion coefficients, $p_1,p_2$, the tangential distortion coefficients, and $r$ the radial distance from the principal point.

\subsubsection{Extrinsic camera parameters:} With the purpose of projecting image-based data accurately to the LiDAR point cloud, we need to find the pose of the camera relative to the LiDAR, i.e., the extrinsic camera matrix parameters, which encode a 6D transformation $\mathbf{T_{l\rightarrow c}}$ between the camera and the LiDAR, in homogeneous matrix form: 

\begin{equation}
\mathbf{T_{l\rightarrow c}}=
\begin{bmatrix}
\mathbf{R} & \mathbf{T}
\end{bmatrix}=
\begin{bmatrix}
r_{xx}     & r_{xy}    & r_{xz}    &t_x\\
r_{yx}     & r_{yy}    & r_{yz}    &t_y\\
r_{zx}     & r_{zy}    & r_{zz}    &t_z\\
0          &    0      & 0         &1
\end{bmatrix}
\end{equation}

\subsection{Camera and LiDAR calibration}
\label{sec:lidar_inertial_calibration}
In order to estimate the intrinsic camera parameters and the LiDAR to camera rotation and translation (i.e. extrinsic parameters), we utilize two widely adopted algorithms, described below.
\begin{figure}
    \centering
    \includegraphics[width=0.9\textwidth]{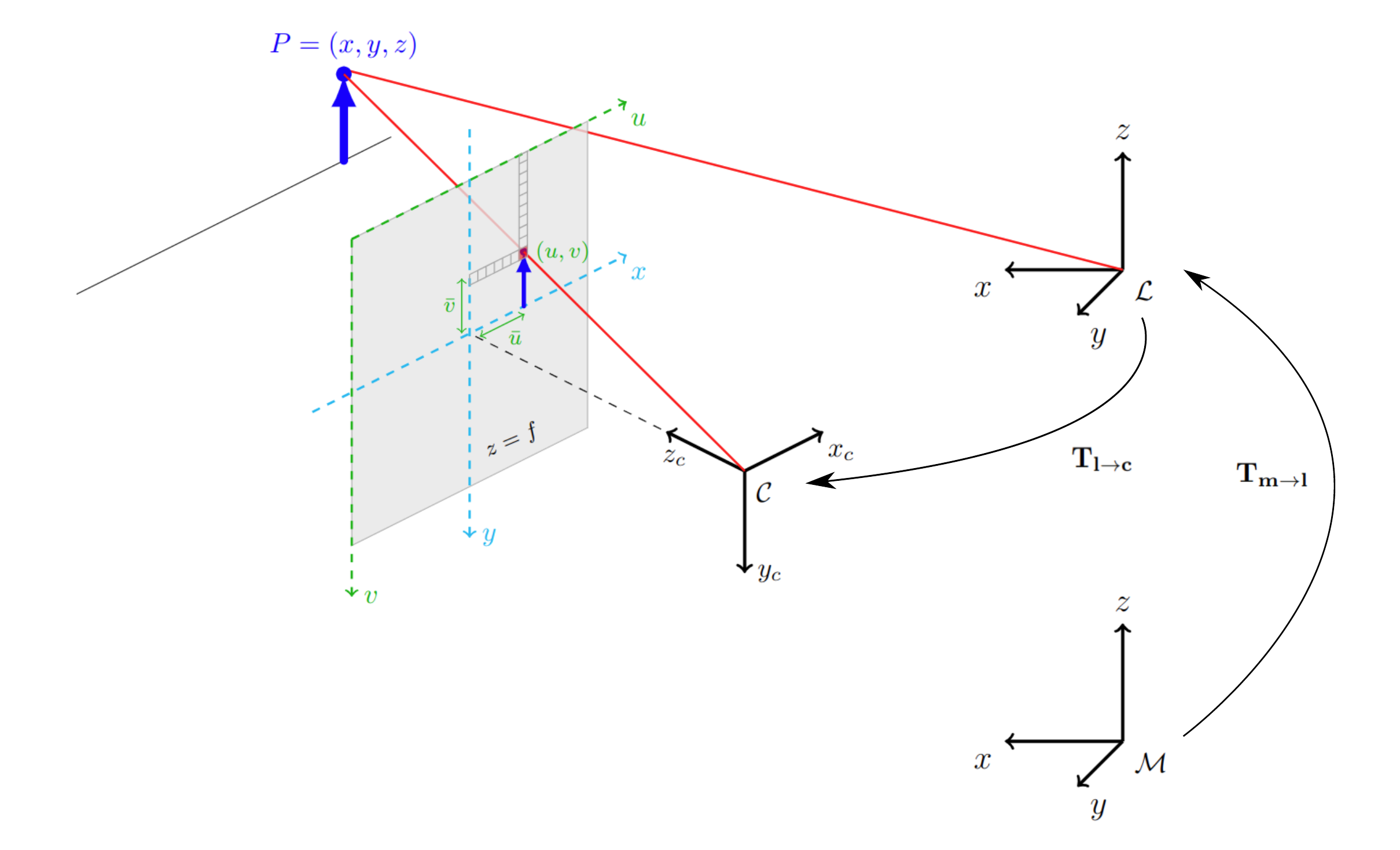}
    \caption{Main coordinate systems and intrinsic parameters considered by our LiDAR camera system, including the camera coordinate system $\mathcal{C}$,  the LiDAR coordinate system  $\mathcal{L}$ and the map coordinate system $\mathcal{M}$}.
    \label{fig:enter-label}
\end{figure}

\subsubsection{Camera intrinsic calibration}

In this work we rely on the camera calibration procedure of \cite{furgale2013unified}, which is a versatile and accurate method for calibrating multiple cameras, and (optionally) multiple IMU sensor suites. The method calibrates both intrinsics and extrinsics and provides a robust parameter identification (i.e. calibration) even in the presence of dynamic motion, in a unified manner. In case of the presence of IMU(s) the procedure incorporates, as well, the IMU biases. The full measurement model includes both camera and IMU measurements, but for the sake of simplicity, we consider only a single camera.

The calibration parameters (intrinsics and extrinsics) are estimated by
capturing multiple images of a calibration checkerboard, with known dimensions, from different viewpoints, i.e., by locating and extracting the image coordinates of the checkerboard corners in each image, using a corner detection algorithm. The 3D points coordinates in world coordinates can be mapped by minimizing the re-projection error $E$, i.e., the discrepancy between the observed image points $(u,v)$ and their corresponding projections $(u',v')$ based on the calibrated camera model, according to 
\begin{align}
E=\sum_i E_i^2 \quad\text{with}\quad E_i =\sqrt{(u'-u)^2+(v'-v)^2}
\end{align}

The parameters are obtained using a nonlinear optimization algorithm (e.g., Levenberg-Marquardt \cite{wright2006numerical}) to minimize the re-projection error. The optimization involves iteratively adjusting both intrinsic parameters and distortion coefficients.

\subsubsection{LiDAR-camera extrinsic calibration}
\begin{figure}
    \centering
    \begin{subfigure}[t]{0.45\textwidth}
        \includegraphics[width=\textwidth]{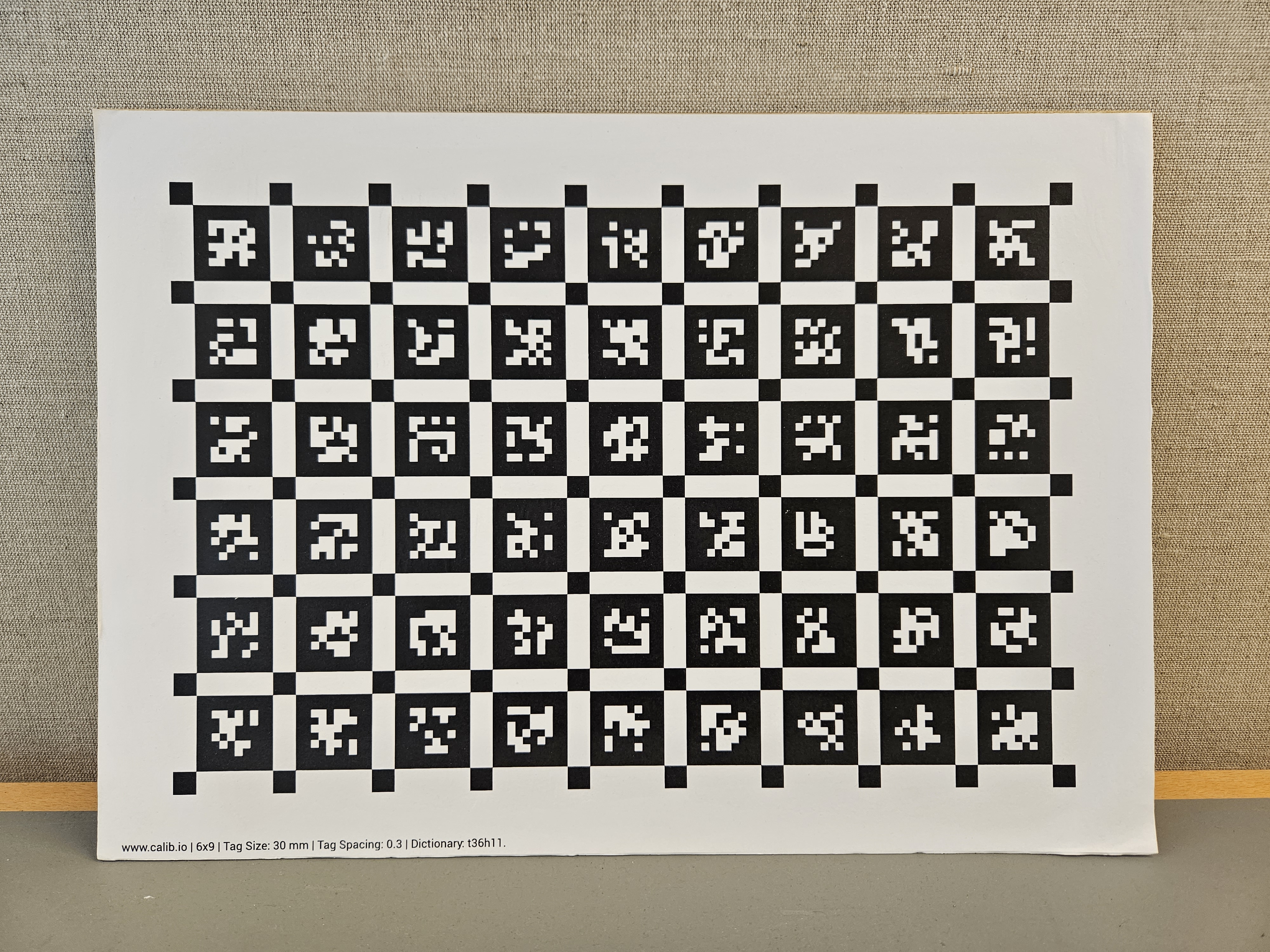}
        \caption{A3-sized AprilGrid camera calibration target, with 8 rows, 12 columns, 22mm tags, and 6.6mm spacing.}
        \label{fig:aprilgrid_target}
    \end{subfigure}
    \begin{subfigure}[t]{0.45\textwidth}
        \includegraphics[width=\textwidth]{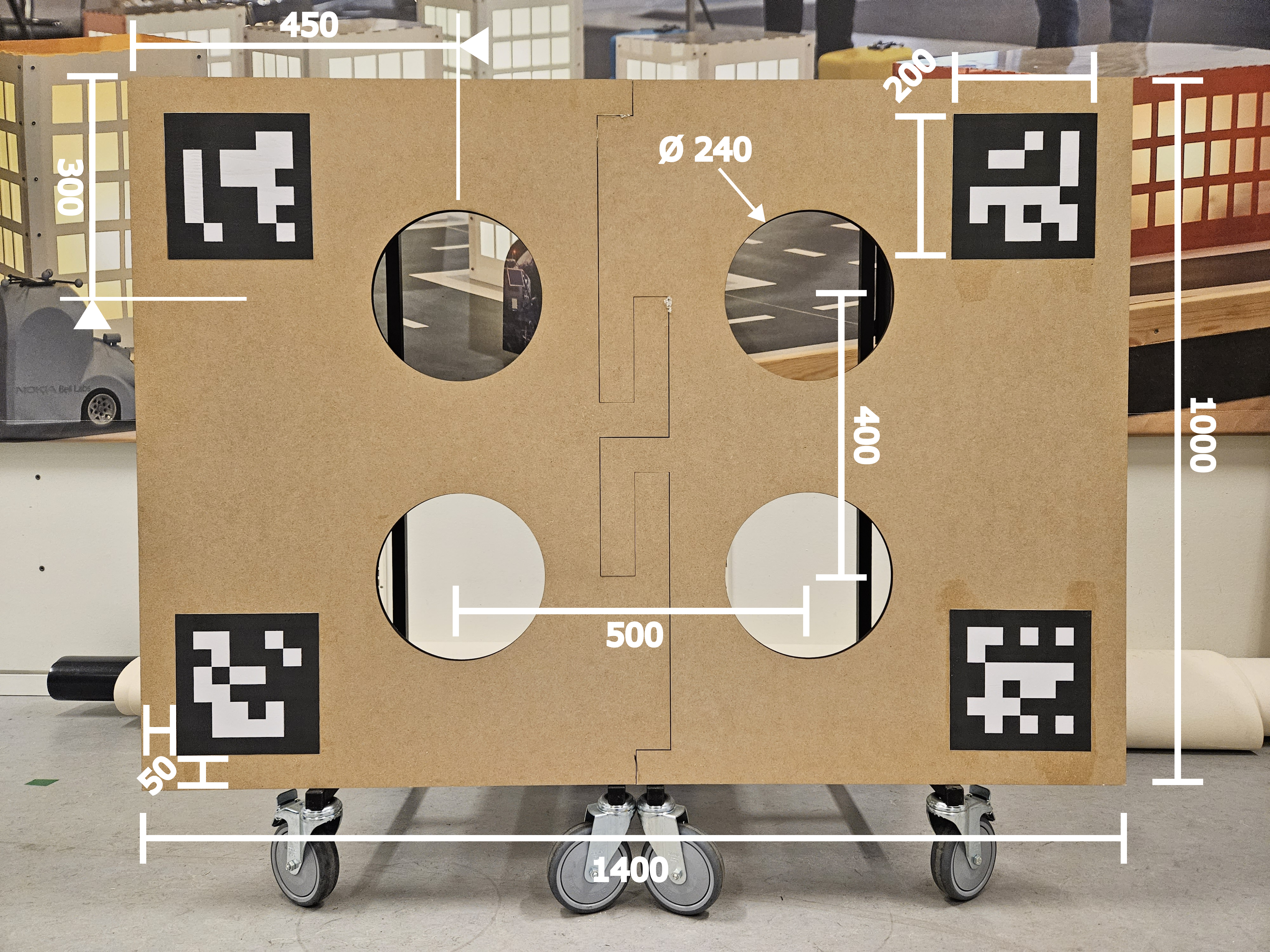}
        \caption{Fabricated extrinsic calibration target based on \cite{Beltran2022Velo2cam}, using the default recommended dimensions. All values are in millimeters.}
        \label{fig:velo2cam_target}
    \end{subfigure}
    \caption{Calibration targets used for intrinsic and extrinsic calibration.}
    \label{fig:calibration_targets}
\end{figure}

To find the relative transformation $\mathbf{T}_{l\rightarrow c}
$ between the LiDAR and camera when sensors are not configured in a predetermined rigid structure, the method described by \citet{Beltran2022Velo2cam} is utilised. It is a two-stage solution for calibrating between pairs of monocular cameras, stereo cameras, and LiDARs, using a custom calibration pattern featuring four circular holes and four ArUco markers \cite{GarridoJurado2014ArUco}, which can be seen in figure \ref{fig:velo2cam_target}.

In short, the method is based on localizing the centroids of the holes in the target across the two sensors being calibrated, then aligning the two sets of 3D reference points.
The following is an overview of the process described by \citet{Beltran2022Velo2cam}. Note that details on calibrating with stereoscopic cameras are briefly covered solely for the sake of completeness.

\paragraph{Reference point estimation}
In the first stage, the goal is to localize the calibration target, and estimate the center points of the holes within, relative to the individual sensors. This is done frame-by-frame, per sensor. Here, there are two pathways: one for 3D input, and one for monocular input.

Stereo-camera feeds are pre-processed into a 3D point cloud, then later used similarly to the LiDAR feed. In both cases, user-adjustable pass-through filters are applied to limit the search space. This requires user intervention, specific to the current physical calibration setup. 

\begin{figure}
    \centering 
    \begin{subfigure}[t]{0.45\textwidth}
        \includegraphics[width=\textwidth]{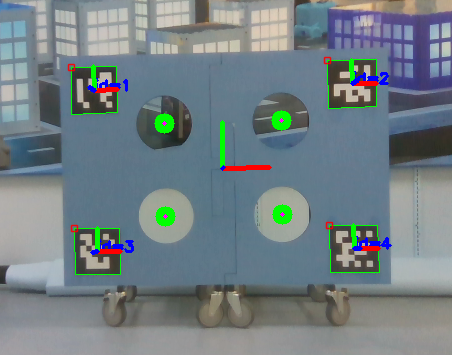}
        \label{fig:ext_calibration_camera_debug}
    \end{subfigure}
    \begin{subfigure}[t]{0.45\textwidth}
        \includegraphics[width=\textwidth]{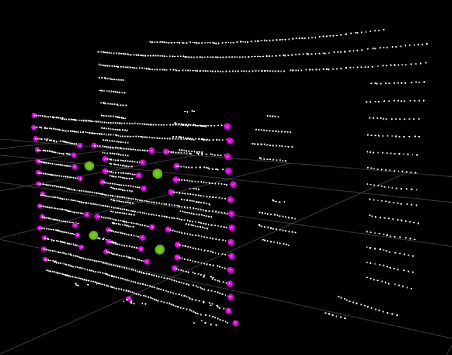}
        \label{fig:ext_calibration_lidar}
    \end{subfigure}
    \caption{Reference point estimation using ArUco markers for monocular cameras (left) and edge detection for LiDARs (right).}
    \label{fig:ext_calibration_refPoints}
\end{figure}

Edge points are then identified. For LiDARs, depth discontinuities in the point cloud are used to detect edges \cite{Levinson2013AutomaticOC}. This assumes that points resulting from LiDAR rays passing through the calibration target holes exist, so care must be taken to not leave excessive open space behind the target, nor filter these points out in the preceding step. A visualization of this can be seen in figure \ref{fig:ext_calibration_refPoints}.
For stereo cameras, a Sobel filter \cite{sobel2014history} is applied to one of the two stereo images, and points are mapped to the resulting pixel values. Points with sufficiently low values are discarded.

It is at this point the two modalities are treated the same. A RANdom Sample Consensus (RANSAC) algorithm \cite{isa2019point} is applied to the filtered point cloud data, in order to fit a plane model, representing the target plane surface. The edge-points from the previous step are filtered using the obtained plane model, such that the resulting point cloud only contains points belonging to the target.

Next, the remaining points are projected on the planar model, resulting in a 2D point cloud. The circular holes are then extracted using a 2D circle segmentation. This is performed iteratively, removing inliers, until the  remaining points are insufficient to describe further circles. At least four circles must be found to proceed, otherwise the current frame is discarded. 
Monocular cameras rely on using the four ArUco markers as an ArUco board to estimate the relative position of the target. An example of this can be seen in figure \ref{fig:ext_calibration_refPoints}.


Once circle centers have been found, the processing of sensor data across modalities are the same. To rule out incorrect detections, a geometric consistency check is performed. The circle center points are grouped in sets of four, and the dimensions of the rectangle they form together are compared against what is theoretically expected. The assumption is that there should only be one valid, geometrically consistent set. If more than one, or zero, valid sets are found, the frame is discarded. Otherwise, the center-points are converted back into 3D space, and the resulting cloud of four points is considered to be valid reference points.

Having a single set of reference points per sensor is technically sufficient, but to reduce potential errors, such as from sensor noise, multiple (i.e. 30) sets are accumulated per sensor, and verified using Euclidean clustering \cite{nguyen20133d}. In case of detecting more than the expected four clusters, the data is considered unreliable, else cluster centroids are used as reference points for the second stage. Furthermore, Beltrán et al \cite{Beltran2022Velo2cam}  allow for (and recommend) accumulating  reference points over \(M\) target poses, as to generate \(4 \times M \) reference points, adding further constraints in order to improve the final results.

\paragraph{Registration procedure}
In the second stage, the goal is to find the rigid transformation that best aligns the reference point sets with each other. This is handled as a multi-objective optimization, involving \(4 \times M \) objective functions. 

First, the reference points must be paired, so that each reference point from one sensor is associated with the corresponding reference point from the other sensor. There is no guarantee that reference points in each set are in the same order, so to ensure correct association, the four reference points per target are converted to polar coordinates, and the top-most point (i.e. lowest inclination) is identified. From hereon, the remaining points can then be identified by comparing distances to them, and thus paired correctly.

\subsection{Image-based corrosion identification}
In this work we rely on the popular Deep Neural Network architecture, U-net \cite{unet2015}, for image-based semantic segmentation. U-net derives from the fully convolutional network introduced in \cite{7298965}, with modifications tailored to facilitate training with small image samples while surpassing its predecessor in accuracy. The architecture of U-net comprises a conventional convolutional network, coupled with two pathways. The contracting path initiates with two 3x3 unpadded convolutions, succeeded by rectified linear units (ReLU) and downsampling through 2x2 max pooling operations.

\begin{figure}[!t]
  \centering   
  \includegraphics[width=0.9\textwidth]{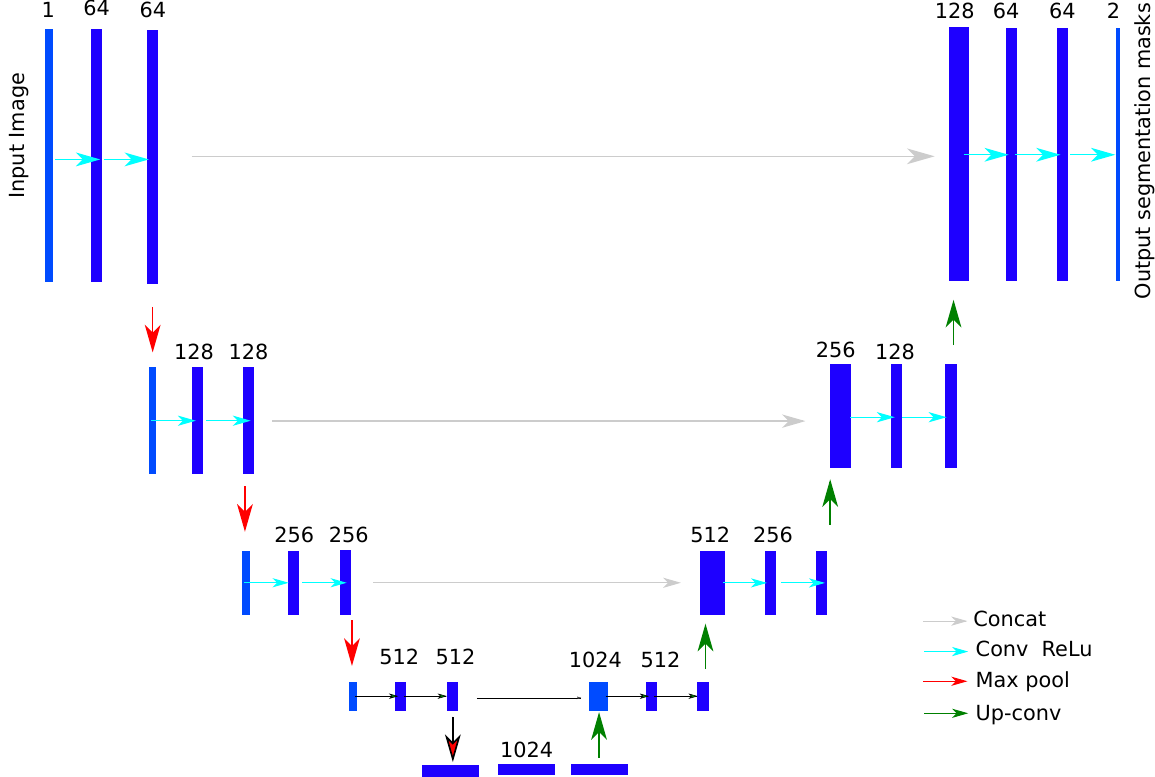}
  \caption{U-Net architecture for a 1024 pixel input (i.e. 32x32) at the lowest resolution. Blue boxes represent multi-channel feature maps, with the number of channels indicated at the top, and the arrows indicate various operations in the network (figure adapted from \cite{unet2015}).}
  \label{fig:unet}
\end{figure}

In the expansive path, the feature map undergoes up-sampling through 2x2 convolutions, followed by cropping to address the loss of border pixels during convolutions. Subsequently, the cropped feature map from the contracting path is concatenated, and 3 × 3 convolutions are applied, followed by rectified linear units (ReLU) as illustrated in Figure \ref{fig:unet}.

U-net's use of convolutional layers with small receptive fields promotes the extraction of intricate hierarchical features, contributing to its robust representation learning capabilities. The architecture's adaptability to limited training data sets it apart, making it suitable for applications where obtaining extensive labeled samples poses challenges.

The symmetric structure of U-net, with its balanced encoding and decoding paths, plays a pivotal role in maintaining spatial relationships, enhancing feature retention, and boosting its overall performance. A clever combination of skip connections, convolutional layers, adaptability to varied data sizes, allows achieving high accuracy with limited training data. 

\subsection{LiDAR-based localization and mapping}
Our pipeline for mapping and localization combines two methods. The first, offline, employs a graph-based SLAM approach to build a point cloud map of the environment. The second, given a map, employs a unscented Kalman filter (UKF) \cite{wan2001unscented} based approach to estimate in real-time the location of the sensor apparatus in the map.


\subsubsection{Graph-based LiDAR SLAM}

LiDAR-based Graph-SLAM is a sensor fusion technique that  utilizes a graph structure to represent the relationships between different poses (positions and orientations) of a LiDAR system over time, along with the associated LiDAR measurements. The optimization process seeks to simultaneously estimate the robot's trajectory and create a consistent map of the environment.

A graph $G=(V,E)$ is built over time, where $V$ is the set of vertices representing sensor poses, and $E$ is the set of edges representing constraints between these poses. The optimal node configuration in the graph, is found using non-linear optimization techniques to minimize the error introduced by the constraints. Let $\mathbf{p}_t$ be the sensors poses at $t$, and $\mathbf{r}_{t,t+1}$ be the relative sensor poses between $t$ and $t+1$ estimated via scan matching \cite{magnusson2007scan}. We add them to the pose graph as nodes $\left[\mathbf{p}_0,...,\mathbf{p}_N\right]$ and edges $\left[\mathbf{r}_{0,1},...,\mathbf{r}_{N-1,N}\right]$.  The graph is optimized using the g2o \cite{g2oframework2011} framework, to build a globally consistent 3D map of the environment. To compensate for accumulated pose errors during mapping, the approach allows incorporating ground plane constraints in the graph pose optimization.

The final output of the algorithm is a map  $\mathcal{M}$ of size $M$, comprising a finite set of points in $\mathbb{R}^3$. Let us denote by $\mathbf{m}_i\in\mathcal{M}, i\in \{1,...,M\}$, each point belonging to $\mathcal{M}$. 

\subsubsection{Real-time LiDAR-based localization}
During the online localization phase, the localization system \cite{koide2019portable} estimates its own pose on the previously created 3D point cloud map, by combining a scan-matching algorithm with an angular velocity-based pose predictor. An UKF \cite{wan2001unscented} is used for sequential Bayesian estimation of the pose of the LiDAR-camera sensor apparatus. The scan matching utilizes a Normal Distributions Transform (NDT) \cite{magnusson2007scan} for estimating the LiDAR motion between consecutive frames, which outperforms other scan matching approaches, such as the iterative closest point (ICP) \cite{icp1992}.

\subsection{Semantic-geometric mapping}

In order to fuse semantic information with the 3D map of the environment, we utilize the known camera intrinsics, extrinsics between the lidar and the camera, and the pose between the LiDAR and the map, given by the localization system.

Given an image $\mathcal{I}$ provided by the system's semantic segmentation algorithm, the semantic color information is projected into the map by combining the transformation between the map and the LiDAR, computed by the localization system $\mathbf{T_{m\rightarrow l}}$, the transformation between the LiDAR and the camera optical center $\mathbf{T_{l\rightarrow c}}$. More specifically, for each point belonging to the map, we compute the corresponding image pixel, using homogeneous coordinates according to,

\begin{equation}
\mathbf{u}_i=
\mathbf{K}{\mathbf{T}}_{l\rightarrow c}{\mathbf{T}}_{c\rightarrow m}
\mathbf{m}_i 
\quad \text{with} \quad
\mathbf{u}_i=
\begin{bmatrix}
u\\
v\\
1\end{bmatrix}
\text{, }
 \mathbf{m}_i=
\begin{bmatrix}
x_{m}\\
y_{m}\\
z_{m}\\
1\end{bmatrix}
\end{equation}
and if the point falls within the image plane, its corresponding color is updated, by averaging its current RGB values with the corresponding image pixel one.

\section{Experiments} \label{sec:experiments}
In this section we describe in detail the selected sensor apparatus as well as indoor and outdoor experiments to assess the performance and applicability of our methods in a real use-case scenario.

\subsection{Hardware design and sensor selection}
The design of the portable sensor-holder was a critical aspect behind our efficient data collection in various applications, such as environmental monitoring or industrial inspections. The sensor-holder is versatile, providing a stable platform for mounting sensors of different sizes and types. In our experiments we utilized an Ouster Rev 06 LiDAR, and a RealSense d435i camera during our experiments. The design was selected to offer ease of mobility to capture diverse environments and facilitate data capture. We selected 3D printed material, to ensure easy adaptations during prototyping (e.g. change of sensors), ensure durability in challenging conditions while being lightweight for comfortable use. An ergonomic design that prioritizes user comfort and ease of handling is crucial for prolonged data collection tasks. Additionally, the sensor-holder incorporates features such as screw holes to facilitate the deployment and interchange of sensors.

Complementing the sensor-holder, the design of a backpack for data collection plays a pivotal role in ensuring seamless mobility and accessibility of necessary equipment. The backpack was engineered to be lightweight, and easy to carry for long-periods of time, distributing the weight evenly, reducing strain on the user during extended periods of data collection. Compartments within the backpack should be strategically arranged to accommodate the sensor-holder, ensuring secure storage and easy access. Integration of a power supply system for sensors, along with cable management solutions, is vital to sustain prolonged data collection sessions. The backpack design should prioritize user comfort with padded shoulder straps and a ventilated back panel to mitigate discomfort caused by extended wear. Furthermore, the incorporation of smart features such as ground truth markers in a controlled laboratory environments allows assessing the overall capabilities of the data collection process, making the backpack an integral component of a well-rounded field data collection system. Please see Table \ref{tab:hardware_selection} for details on the selected hardware.




\begin{table}[]
\scriptsize
\centering
\caption{Hardware used in the experiments.}
\label{tab:hardware_selection}
\begin{tabular}{ll}
\hline
Hardware                                                     & Used in            \\ \hline
Ouster OS1-32 Rev 06 LiDAR with uniform beam spacing         & All experiments    \\
Intel Realsense D435i depth camera                           & All experiments    \\
Intel NUC 12 Pro computer with Intel Core i7-1260P processor & All experiments    \\
Portable backpack power system with 4 lithium batteries      & All experiments    \\
OptiTrack motion capture system with dedicated computer      & Indoor experiments \\
High definition webcam with dedicated computer               & Indoor experiments \\
Gigabit ethernet switch                                      & Indoor experiments \\ \hline
\end{tabular}
\end{table}

\subsection{Indoor experimental setup}

In order to test the proposed system, and compare our solutions for LiDAR-based localization, experimental data recording sessions in an indoor laboratory setting were conducted. The purpose for these recordings were to create datasets for assessing the performance of the localization algorithms, using a motion capture system. Furthermore, uncorroded and corroded metal parts were placed in the environment, to allow testing the performance of the corrosion detection algorithms. 

The setup was as follows: LiDAR and depth camera sensors were affixed to the custom 3D print with handles, as shown in Figure \ref{fig:backpac-system}. 
The NUC 12 computer, along with LiDAR connector box, were installed into and powered by the backpack. Prior to recording, the motion capture system was calibrated, and the tracked area was marked with tape. A rigid-body tracking marker was attached to the top of the LiDAR.

\begin{figure}[H]
    \centering
    \includegraphics[width=0.96\linewidth]{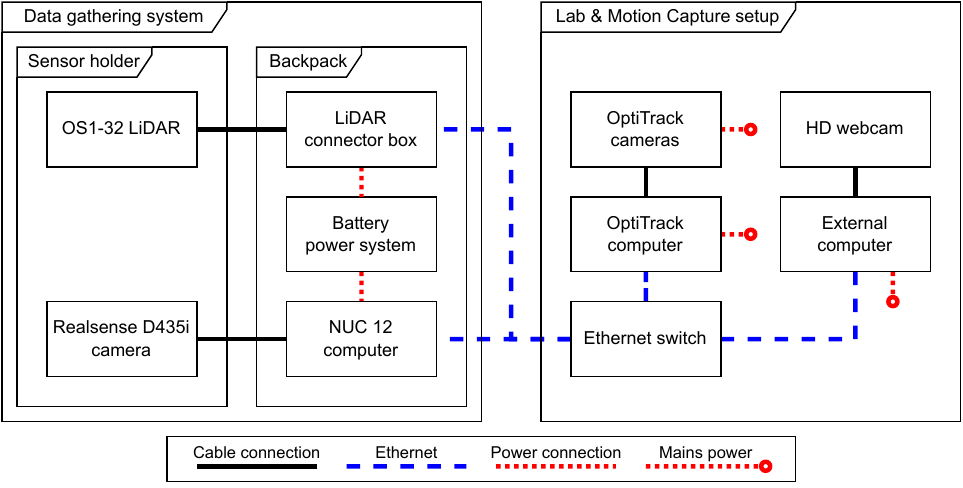}
    \caption{Hardware setup. The Lab \& Motion Capture setup is exclusively used in indoor experiments. }
    \label{fig:system_diagram_complete}
\end{figure}


\begin{figure}[H]
    \centering
    \begin{subfigure}[t]{0.487\textwidth}
        \centering
        \includegraphics[width=\textwidth]{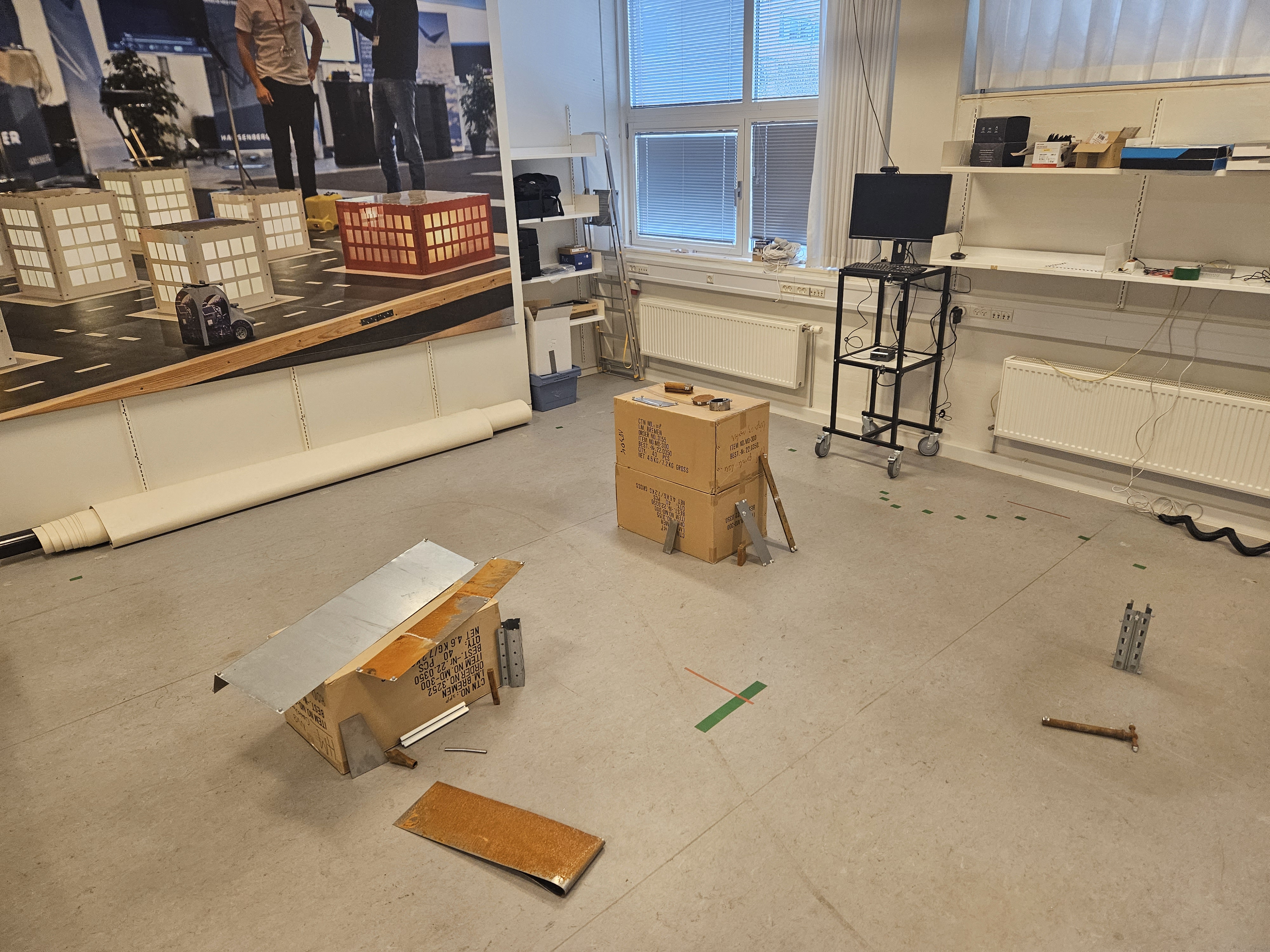}
        \caption{General view of the indoor experimental area.}
        \label{fig:tracked_area}
    \end{subfigure}
    \begin{subfigure}[t]{0.487\textwidth}
        \centering
        \includegraphics[width=\textwidth]{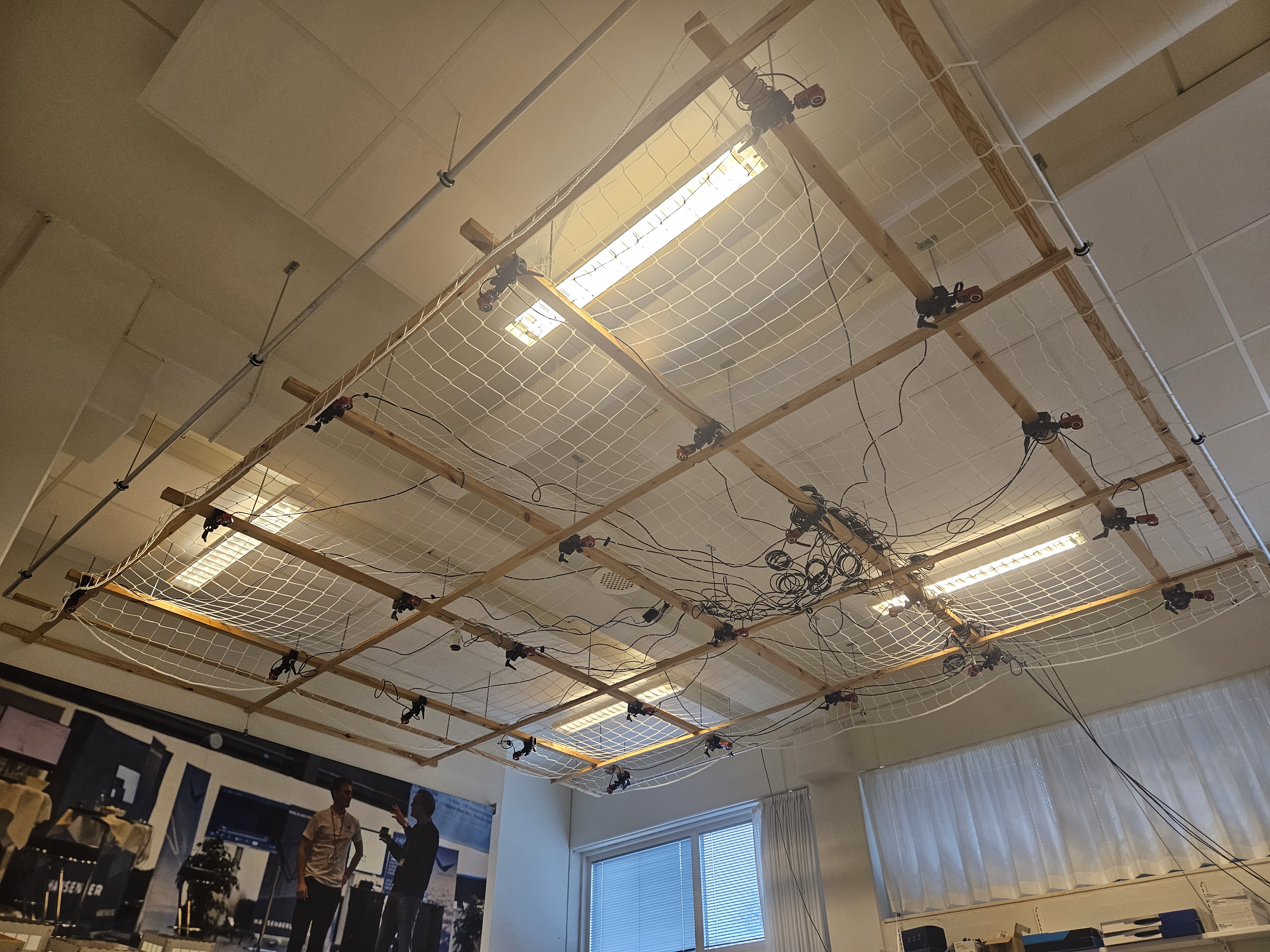}
        \caption{Motion capture system installed on the lab ceiling.}
    \end{subfigure}
\caption{Pictures from indoor experimental setup.}
\label{fig:tracking_system}
\end{figure}

\subsubsection{Localization and mapping performance}


In order to assess the performance of the indoor localization and mapping algorithms, we integrated a high precision optitrack motion capture system, with $0.1mm$ precision, for ground truth acquisition \cite{furtado2019comparative}, installed in the ceiling of the lab (see Figure \ref{fig:tracking_system}). A set of markers were placed on top of the LiDAR-camera system in order to build a trackable reference frame. To quantitatively evaluate the performance of the algorithms, we compared the trajectory given by the ground truth system with the computed trajectory UKF localization algorithms.
More specifically, trajectories were temporally and spatially aligned using the Umeyama alignment algorithm \cite{kabsch1976solution}, and the absolute and relative pose errors were used for quantitative evaluation, i.e., to numerically assess the consistency of the trajectory. 

While the absolute pose error (APE) focuses on the accuracy of the overall position and orientation of the LiDAR-camera system in the ground truth frame, i.e., global trajectory consistency, the relative pose error (RPE) is concerned with the accuracy of the changes in pose between consecutive measurements in a local coordinate frame, hence suitable for measuring drift in odometry systems.

\begin{figure}[H]
    \centering
    \begin{subfigure}{0.45\textwidth}
        \includegraphics[width=0.99\textwidth]{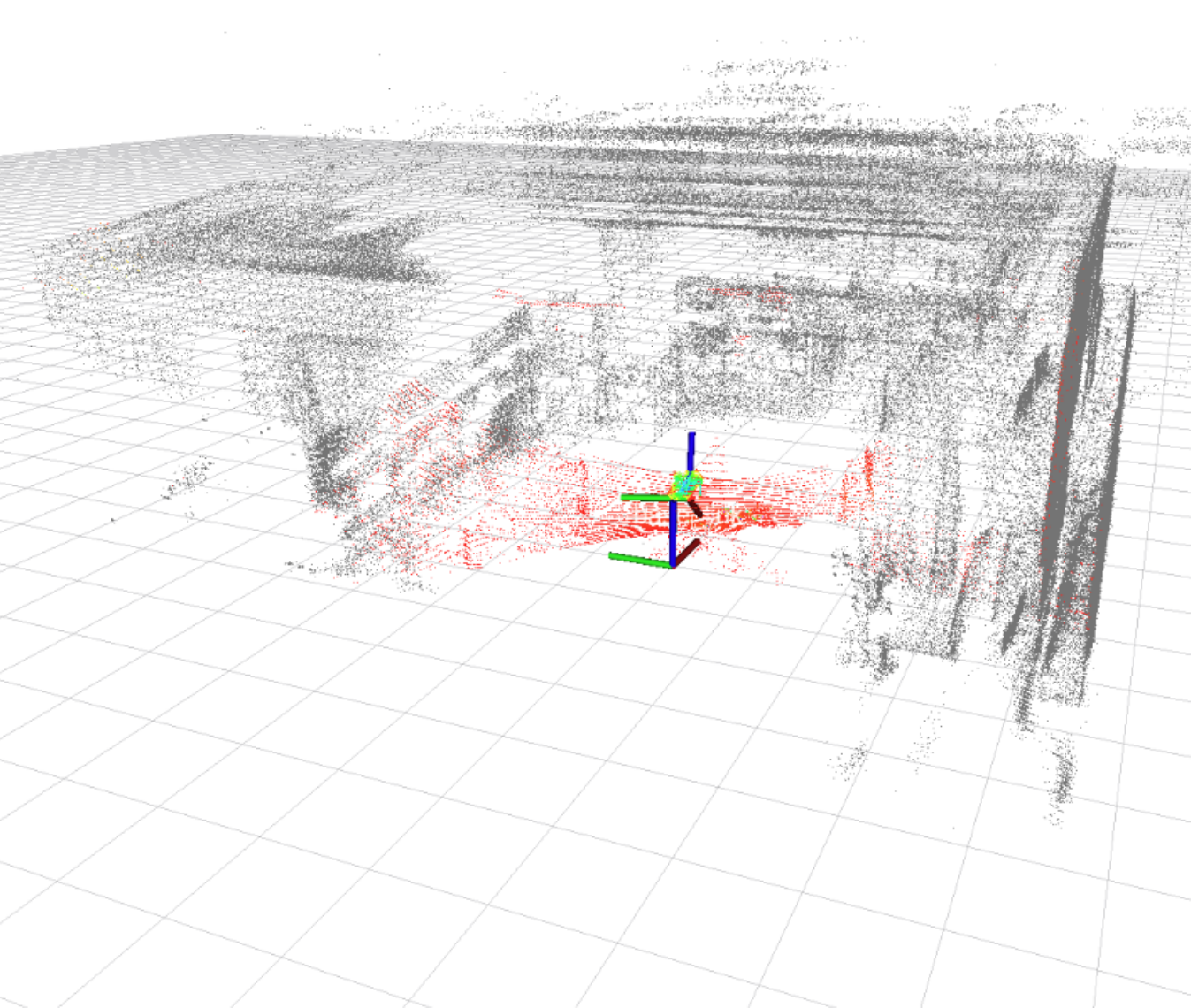}
        \caption{Map and LiDAR frames} 
    \end{subfigure}
    \begin{subfigure}{0.45\textwidth}
        \includegraphics[width=0.99\textwidth]{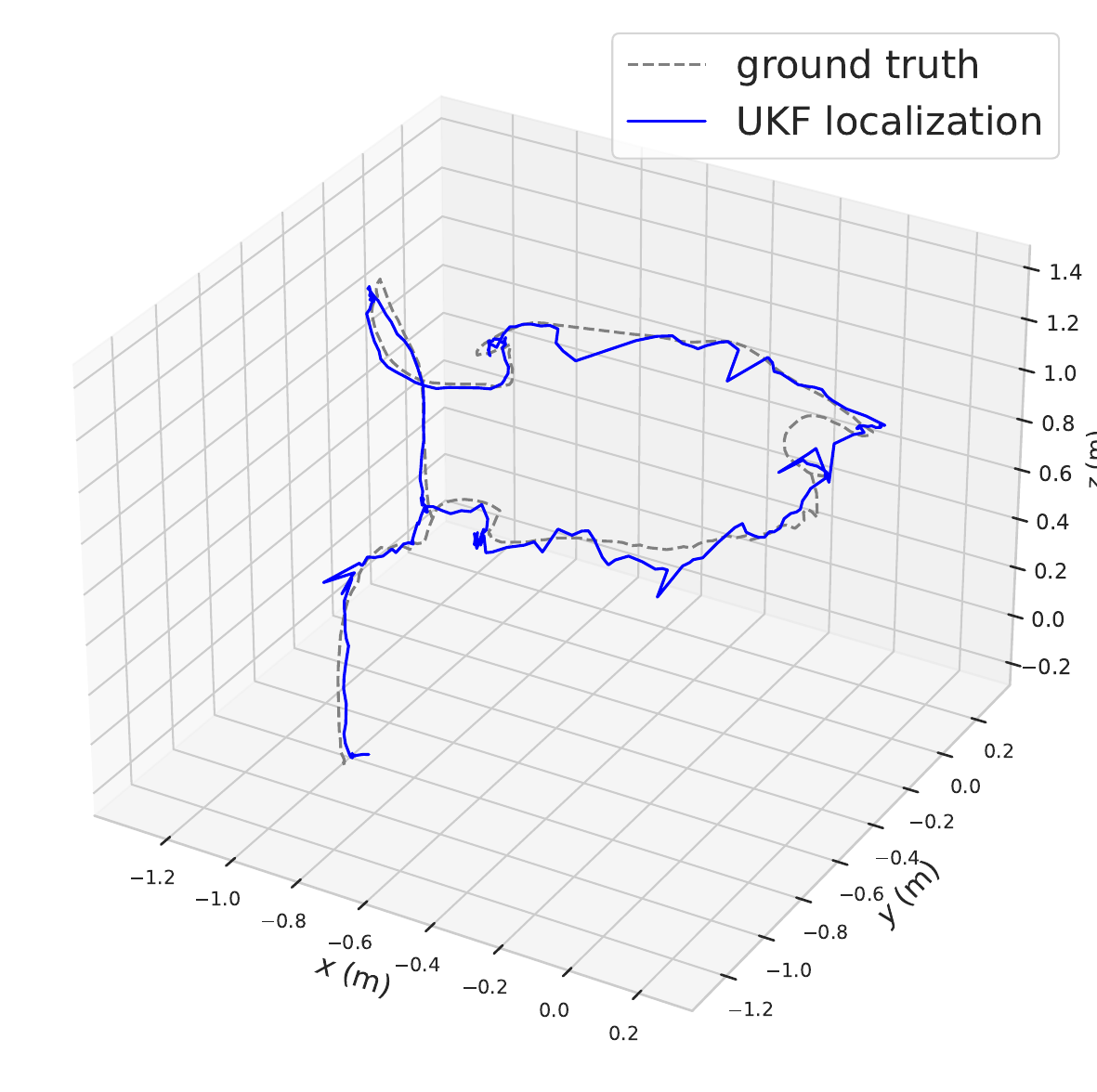}
        \caption{3D trajectories} 
    \end{subfigure}
    \begin{subfigure}{0.49\textwidth}
        \includegraphics[width=0.99\textwidth]{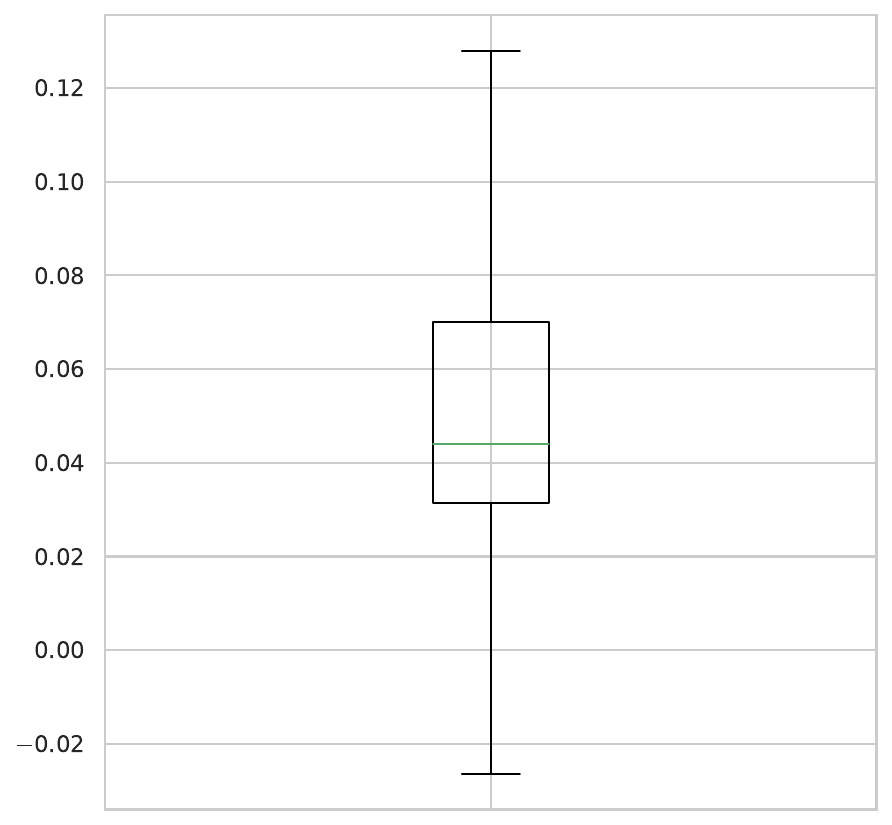}
        \caption{Absolute Pose Error (m)} 
    \end{subfigure}
    \begin{subfigure}{0.49\textwidth}
    \includegraphics[width=0.99\textwidth]{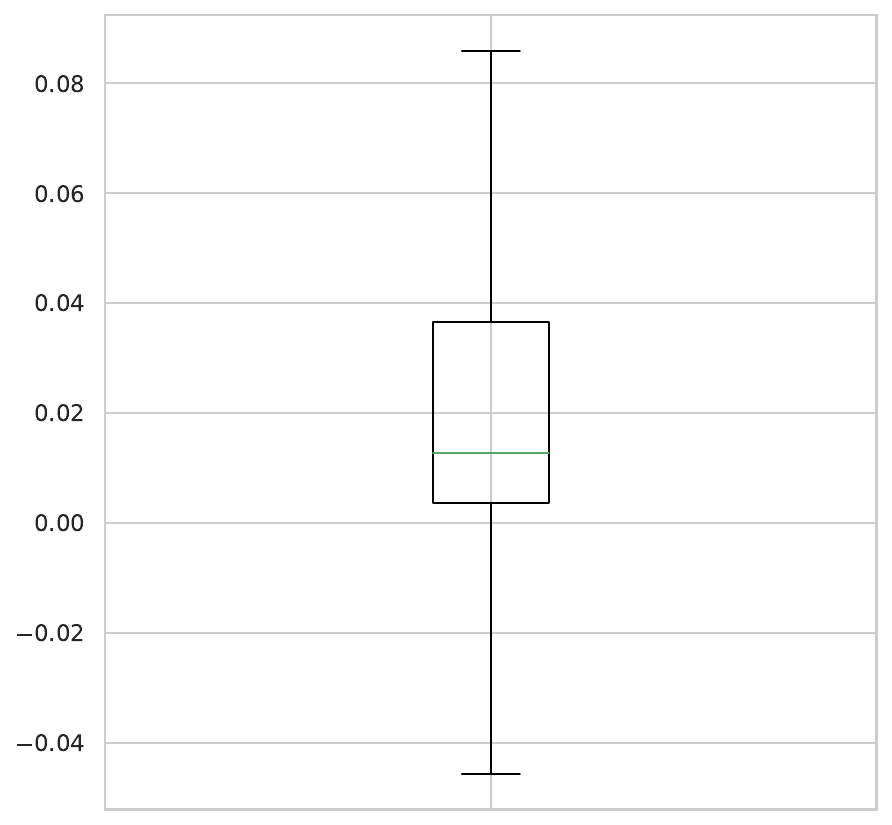}
    \caption{Relative Pose Error (m)} 
    \end{subfigure}
    \caption{LiDAR-based localization performance in indoor experimental scenario.} 
\label{fig:localization_accuracy_indoor} 
\end{figure}

Figure \ref{fig:localization_accuracy_indoor} (a) depicts the map point cloud obtained with the SLAM algorithm, and  the map and LiDAR reference frames during online localization. Figure  \ref{fig:localization_accuracy_indoor} (b), (c) and (d) show the 3D trajectories of the ground truth and UKF localization algorithm, and the absolute and relative pose errors, respectively. The results demonstrate the high-accuracy of the used LiDAR-based localization system with less then $0.05m$ and $0.02m$ average absolute and relative position errors, respectively. One setback, though, of the employed method resides on the fact that it needs a good pose initialization. However, this limitation can be easily surpassed with the use of Monte Carlo Localization (MCL) based methods (e.g. \cite{bedkowski2017online}), at the cost of increased computation effort.

\subsection{Semantic segmentation}

In this section, we conduct a comparative analysis of advanced semantic segmentation networks outlined in the preceding section, specifically focusing on their effectiveness in segmenting corrosion in metallic structures. Throughout our experiments, we employed a 12th Gen Intel® Core™ i9-12900KF processor (24 cores) and a GeForce RTX 3090ti graphics card for both training and testing.

The dataset utilized for training and testing comprises a total of 14,265 labeled images. These images were captured in a high-definition offshore environment using a DSLR camera and were manually annotated using an online labeling tool \cite{segments2022}.

In our experimental approach, we trained a U-net semantic segmentation model, using random crops of size $1024 \times 1024$ extracted from the original input images. The training batch size was set to 8. The dataset was partitioned into training (60\%), validation (20\%), and testing (20\%) sample sizes.  The model underwent pre-training on the ImageNet \cite{krizhevsky2017imagenet} dataset. We use a ResNet-34 as our U-net backbone \cite{koonce2021resnet}, that consists of 34 layers, being popular in image recognition tasks, due to its clever balance between depth and computational efficiency.

\begin{figure*}[!t]
 \centering
 \begin{subfigure}{0.99\textwidth}
 \includegraphics[width=0.99\textwidth]{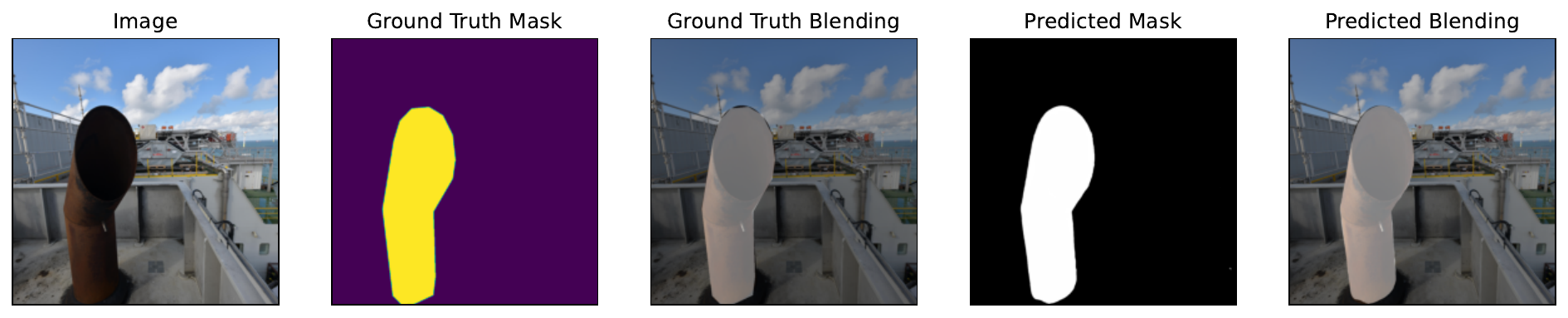}
 \end{subfigure}
  \begin{subfigure}{0.99\textwidth}
  \includegraphics[width=0.99\textwidth]{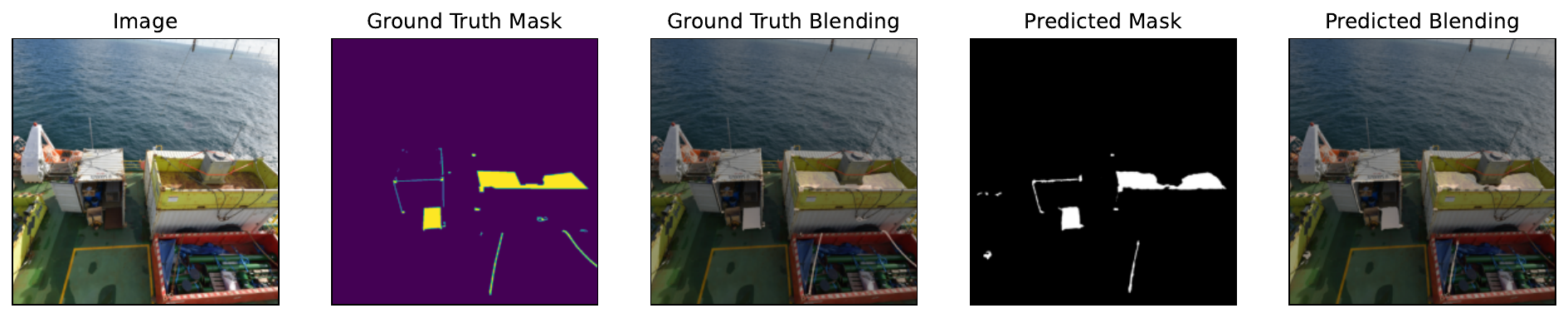}
 \end{subfigure}
 \begin{subfigure}{0.99\textwidth}
  \includegraphics[width=0.99\textwidth]{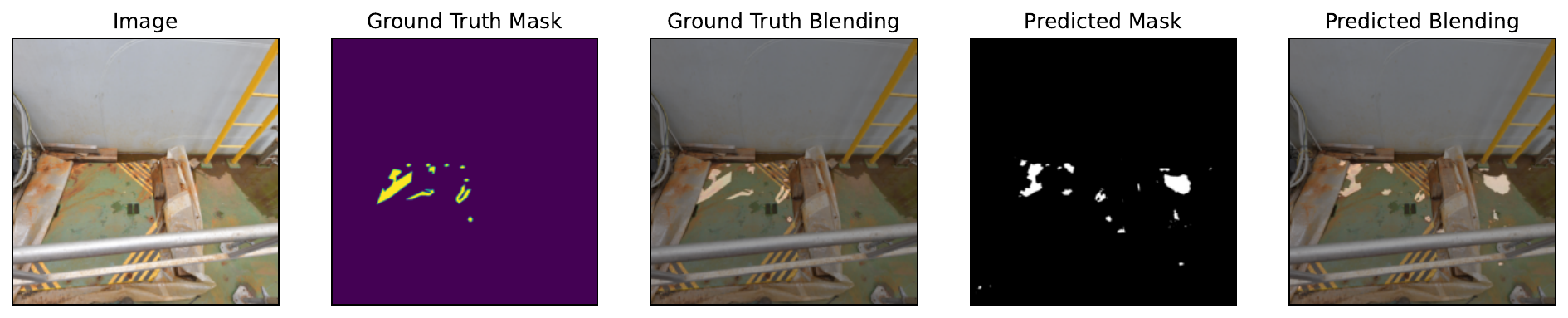}
\end{subfigure}
 \begin{subfigure}{0.99\textwidth}
     \includegraphics[width=0.99\textwidth]{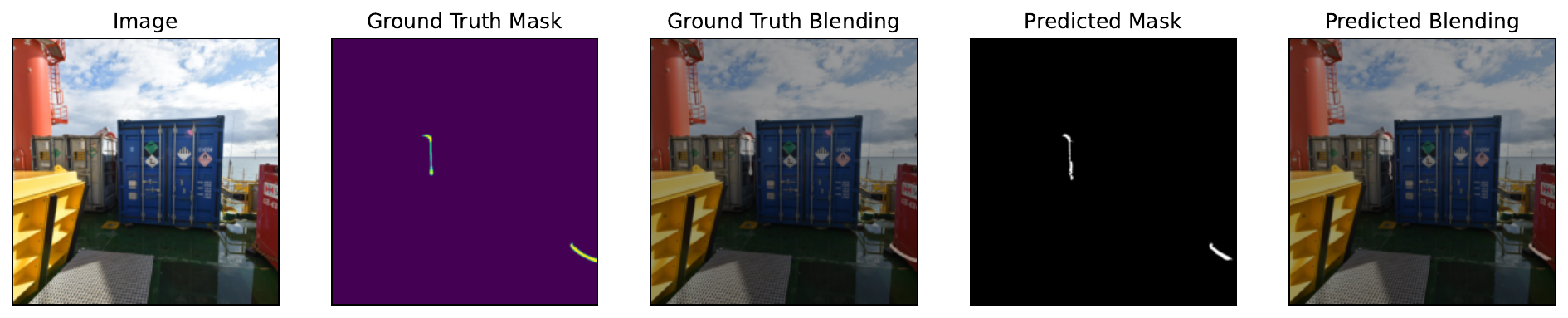}
 \end{subfigure}
 \caption{Semantic segmentation masks generated for different example scenes utilizing the U-net architecture featuring a ResNet34 backbone.} 
\label{fig:semantic_segmentation_unet_scenes} 
\end{figure*}

Figure \ref{fig:semantic_segmentation_unet_scenes} shows qualitatively, the performance of U-net on different scenes. U-net is capable of identifying most ground truth spots, while correctly identifying corrosion spots missed by the labeler. Given the difficulty of accurately labeling corrosion, especially when small spots are easily overlooked by human annotators, precision may not be the most suitable metric for evaluating task effectiveness in this context. As quantitatively shown in Table \ref{table:quantitative_results}, the method exhibits fast average inference times ($0.0445s$), being suitable for real time application. Although our model achieves high precision when compared with similar corrosion identification approaches, e.g. \cite{doi:10.1061/(ASCE)CP.1943-5487.0001045}, being able to recognize most ground truth corrosion pixels, without many false positives, recall is relatively low, due to a high rate of false negatives, i.e., missed corrosion spots in the ground truth. We hypothesize that this is due to a relatively high number of missed corrosion spots by the labeler.

\begin{table*}[t]
\caption{U-net performance results on corrosion segmentation in offshore environments.}
\centering
\label{table:quantitative_results}
\scriptsize
\begin{tabular}{lrrrrrrrlrr}
 & IoU score & Precision &  Recall & F-score &  Avg. inference time (s) \\
model &  &  &  &  &  &  &  &  \\
UNET-resnet34      & 0.4519 & 0.7112 & 0.5508 & 0.5574 & 0.0445 \\
\end{tabular}
\end{table*}

\begin{table*}[]

\centering
\caption{Dataset used for training and validating the semantic segmentation networks.}
\begin{tabular}{l|lll}
                          & \multicolumn{3}{l}{total images in dataset} \\ \hline
train                     & \multicolumn{3}{l}{8559 (60\%)}             \\
\multicolumn{1}{c|}{val}  & \multicolumn{3}{l}{2853 (20\%)}             \\
\multicolumn{1}{c|}{test} & \multicolumn{3}{l}{2853 (20\%)}                       
\end{tabular}
\end{table*}

\section{Conclusions and future work} \label{sec:conclusion}
In this work, we proposed a complete system for corrosion identification and 3D mapping of industrial environments. The proposed designed system comprises a portable sensor-holder for a 3D LiDAR and an RGB camera. The system leverages the accuracy of LiDAR data for 3D localization and mapping, with RGB camera information for semantic segmentation of corroded metal structures. Our proposed framework allows building a 3D geometric-semantic map, by fusing semantic data with a prior known 3D map, via known camera and lidar calibration parameters, and a real-time UKF LiDAR-based localization system.

A set of indoor experiments, assessed the accuracy of the individual parts of the mapping and localization system in a laboratory environment. 

For future work we intend to replace the current UKF-based localization system with a monte-carlo one, to avoid the need of knowing in advance, the approximate initial location of the system. Also, we intend to improve our semantic map representation, with one that allows fusing semantic and geometric data over time in a probabilistic fashion (e.g. using a probabilistic semantic occupancy grid). Ultimately, we plan to enhance our existing semantic segmentation algorithm by training with large amounts of synthetically generated data.

\section*{Acknowledgement}
This work was supported in part by the Danish Energy Agency through the project  
"Predictive Automatic Corrosion Management" (EUDP 2021-II PACMAN, project no.: 64021-2072); and in part by the Spanish Ministry of Science and Innovation under Ramon y Cajal Fellowship number RYC-2020-030676-I funded by MCIN/AEI /10.13039/501100011033 and by the European Social Fund ``Investing in your future''. The authors would further like to thank Semco Maritime for bringing up use-case challenges, and to Martin Bieber Jensen for designing and 3D printing the sensor suite holder.
\bibliographystyle{elsarticle-num-names} 
\bibliography{bibliography.bib}

\end{document}